\documentclass{article}

\usepackage{microtype}
\usepackage{graphicx}
\usepackage{subcaption}
\usepackage{booktabs} 

\usepackage[most]{tcolorbox}
\usepackage{courier} 

\usepackage{hyperref}

\usepackage[accepted]{icml2026}

\usepackage{amsmath}
\usepackage{amssymb}
\usepackage{mathtools}
\usepackage{amsthm}

\usepackage{float}
\usepackage{multirow}
\usepackage{colortbl}

\usepackage[capitalize,noabbrev]{cleveref}

\theoremstyle{plain}

\theoremstyle{definition}

\theoremstyle{remark}

\newcommand\vwm{DynaVieW}

\newcommand{\ph}[1]{\textless #1\textgreater} 
\usepackage{amsmath}
\usepackage{xcolor}
\newcommand{\meanstd}[2]{
  #1\scriptsize\,($\pm$ #2)
}

\definecolor{myblue}{RGB}{92, 147, 195}
\definecolor{mygreen}{RGB}{91, 169, 59}
\definecolor{myorange}{RGB}{200, 122, 67}

\usepackage[textsize=tiny]{todonotes}

\icmltitlerunning{DynaVieW: Schema-Guided World Modeling for Understanding Hierarchical Visual Dynamics}

\begin{document}

\twocolumn[
  \icmltitle{DynaVieW: Schema-Guided World Modeling for \\ Understanding Hierarchical Visual Dynamics}



  \icmlsetsymbol{equal}{*}

  \begin{icmlauthorlist}
    \icmlauthor{Silin Gao}{epfl,equal}
    \icmlauthor{Hao Zhao}{epfl,equal}
    \icmlauthor{Zeming Chen}{epfl}
    \icmlauthor{Sepideh Mamooler}{epfl}
    \icmlauthor{Antara Raaghavi Bhattacharya}{epfl,harvard} \\
    \icmlauthor{Qiyu Wu}{sony}
    \icmlauthor{Hiromi Wakaki}{sony}
    \icmlauthor{Yuki Mitsufuji}{sony}
    \icmlauthor{Li Mi}{epfl,ethz}
    \icmlauthor{Syrielle Montariol}{epfl}
    \icmlauthor{Antoine Bosselut}{epfl}
  \end{icmlauthorlist}

  \icmlaffiliation{epfl}{EPFL, Switzerland}
  \icmlaffiliation{harvard}{Harvard University, United States}
  \icmlaffiliation{sony}{Sony Group Corporation, Japan}
  \icmlaffiliation{ethz}{ETH Zurich, Switzerland}

  \icmlcorrespondingauthor{Silin Gao}{silin.gao@epfl.ch}
  \icmlcorrespondingauthor{Hao Zhao}{hao.zhao@epfl.ch}
  \icmlcorrespondingauthor{Antoine Bosselut}{antoine.bosselut@epfl.ch}

  \icmlkeywords{World Modeling, Visual Dynamic Understanding, Multimodality, Schema Learning}

  \vskip 0.3in
]



\printAffiliationsAndNotice{\icmlEqualContribution}

\begin{abstract}
  Multimodal LLMs struggle to systematically model the temporal evolution of visual scenes in videos or multi-image sequences. Such inputs require models to predict or simulate multiple levels of dynamic constituents, such as actions taken in the visual sequence, and the associated changes to the visual environment that result.
  To address this challenge, we propose a dynamic schema-guided world model, DynaVieW, optimized for visual dynamic prediction and simulation.
  DynaVieW achieves an in-depth understanding of visual dynamics by learning interleaved state-transition sequences, where states cover broad visual scenes from video keyframes, and transitions capture comprehensive dynamic constituents within a hierarchical schema.
  DynaVieW jointly models transition prediction and state simulation under a mixture-of-experts architecture, with a cross-expert selective attention and a schema token re-weighted loss, to ensure effective and robust learning.
  DynaVieW's understanding of visual dynamics boosts its downstream performance in visual narrative creation and world simulation, showing improved consistency, controllability, and instruction-following.\footnote{We release our data and code to the community, our project GitHub: \url{https://github.com/Silin159/DynaVieW}}
\end{abstract}

\section{Introduction}

Large language models (LLMs) are trained for increasingly advanced foundational capabilities \citep{grattafiori2024llama,singh2025openai,comanici2025gemini,yang2025qwen3}. No longer purely limited to natural language processing, they perform multimodal understanding and generation, enabling abilities such as world perception and simulation \citep{qin2024worldsimbench,duan2025worldscore}, and tasks such as visual narrative creation \citep{gao2025vinabench,zhuang2025vistorybench}.

\begin{figure}[t]
\centering
\includegraphics[width=1.0\columnwidth]{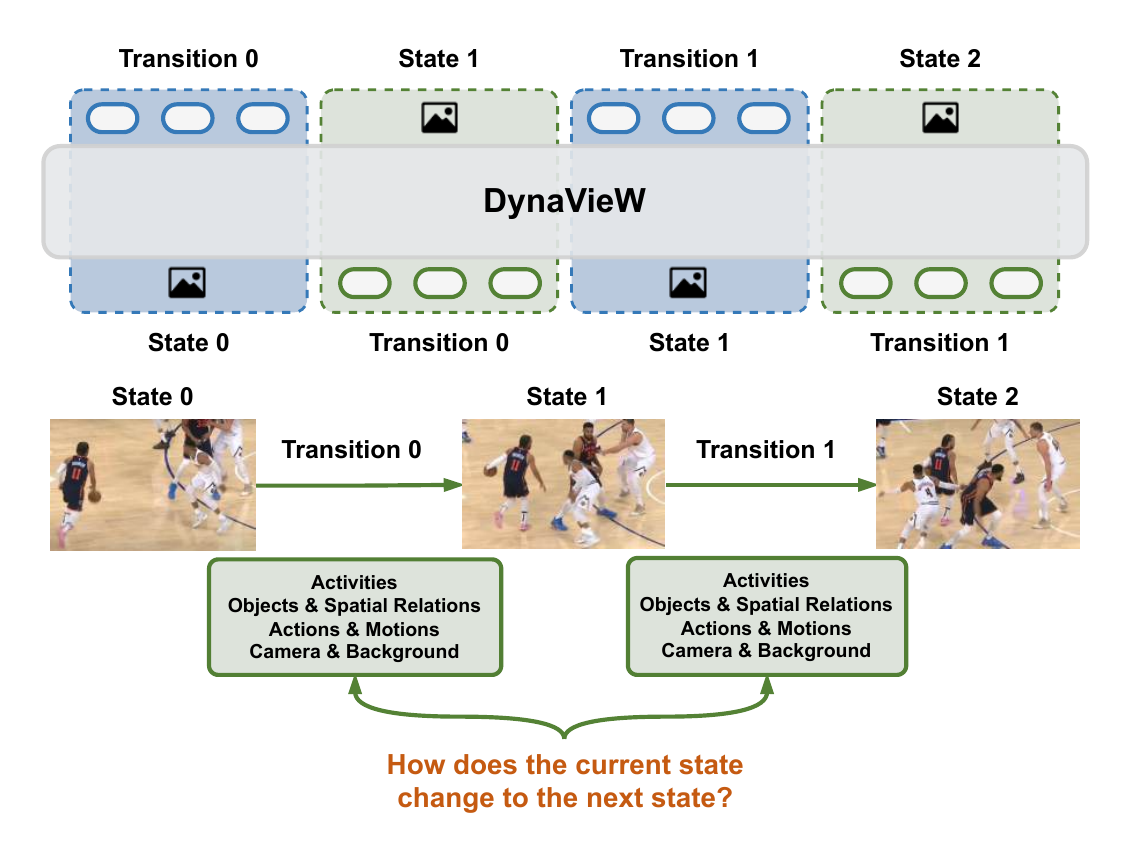}
\caption{\textbf{Overview of \vwm{}.} We formulate our world modeling task as learning interleaved state-transition sequences. In particular, \vwm{} jointly learns the prediction of hierarchical transition descriptions (\textcolor{myblue}{blue}) and the simulation of visual states (\textcolor{mygreen}{green}) in an alternating manner.}
\label{overview_dynaview}
\end{figure}

Despite their improvements on static multimodal tasks (\textit{e.g.}, image captioning), multimodal LLMs still struggle to understand \textit{visual dynamics} --- the changes that occur in visual scenes as actions are carried out. 
For example, a visual scene of \textit{people playing basketball} consists of various \textit{dribbling} and \textit{passing} actions, which further include fine-grained \textit{body movements} and \textit{changes of spatial relations between players}, as shown in Figure~\ref{overview_dynaview}.
Simulating such a scene requires multimodal LLMs to understand these multi-level \textit{constituents} (e.g., high-level events such as actions, low-level transformations in the scene linked to these actions) and how they change the progression of the scene. 
Failing to learn such visual dynamics makes MLLMs unreliable, \textit{e.g.}, they may be inconsistent when generating longer-form visual sequences \citep{gao2025vinabench,zhuang2025vistorybench}, and uncontrollable, as natural language prompts may not align to robust visual understanding \citep{wu2024conceptmix}.

To improve visual dynamic understanding and generation, prior work pre-trains multimodal LLMs on interleaved vision-text data \citep{alayrac2022flamingo,zhu2023multimodal}, which uses natural sequences of visual content, such as storyboard images or video keyframes, that natively contain rich visual dynamics.
In these datasets, the text between visual inputs enables in-place explanations of each step or key point in the progression of the visual narrative, which promotes the understanding of visual dynamics and more precise control of visual sequence simulation.
Despite these datasets, most prior interleaved vision-language LMs still fall short of systematically learning visual dynamics, with many approaches \citep{tian2024mm,chern2024anole,lin2024learning} exposed to pre-training data that only weakly aligns the visual and textual components of interleaved sequences. Other methods were trained on only limited varieties of visual dynamics, such as motion and scene shifts \citep{deng2025emerging}, robot-view actions \citep{qu2025eo}, or game-view interactions \citep{zhang2025matrix,he2025matrix}.

In this paper, we elevate the world modeling capabilities of interleaved vision-language LMs via continued pre-training on broader and more comprehensive descriptions of visual dynamics.
Our proposed \textbf{Dyna}mic \textbf{Vi}sual Sch\textbf{e}ma-guided \textbf{W}orld model (\textbf{\vwm{}}) is pre-trained on interleaved state-transition sequences, as shown in Figure~\ref{overview_dynaview}, across broad domains.
Specifically, \vwm{} learns to simulate visual state sequences that are sourced from keyframes of diverse real-world videos, covering various human daily activities, robotic manipulations, creative works, and auto-driving recordings. 
\vwm{} is trained to predict \textit{transitions} between visual states, represented as text formatted in a hierarchical JSON schema, to structurally capture multiple granularities of transition elements, such as the high-level progression of activities, and the low-level spatial details.
\vwm{} adopts a mixture-of-Transformer-experts (MoT) architecture \citep{deng2025emerging} with a shared multimodal selective attention that facilitates robustly learning long state-transition sequences by dropping out redundant historical information.
Finally, \vwm{} uses a schema token re-weighted cross-entropy (CE) loss to balance its learning of the transition schema format and more specific slot values filled into the schema.

Our \vwm{} model, trained on our broad state-transition data, reaches a more in-depth understanding of visual dynamics in the world, enabling greater controllability on visual sequence generation by verbalizing hierarchically-defined transition characteristics.
In a visual narrative \citep{gao2025vinabench} task, narrative images generated by \vwm{} achieve significantly better consistency and overall quality compared to baseline models, no matter whether prompted for zero-shot generalization or fine-tuned on task-specific data.
DynaView's visual narrative generation is also more controllable, adapting outputs more flexibly when prompted with varying low-level scene descriptions even when the high-level storyline remains the same. 
Finally, we show \vwm{} outperforms baseline models on world simulation tasks \citep{lai2024lego}, demonstrating better instruction following in more deterministic visual generation scenarios.

\section{World Modeling on Hierarchical Dynamics}
We formulate our \vwm{} modeling task as learning interleaved state-transition sequences, as shown in Figure~\ref{overview_dynaview}.
The states in each sequence are images depicting a dynamically changing visual scene, such \textit{a person playing the keyboard} in Figure~\ref{data_construction}, and the transitions are JSON-schema texts hierarchically describing changes between adjacent states, such as \textit{the player's right hand moving down to touch the keyboard}.
\vwm{} jointly learns the prediction of each transition and the simulation of each state (excluding the first initial state) based on preceding states and transitions, which fosters in-depth understanding of visual dynamics.

\subsection{\vwm{} State-Transition Data Construction}
\label{sec_data}
\begin{figure}[t]
\centering
\includegraphics[width=1.0\columnwidth]{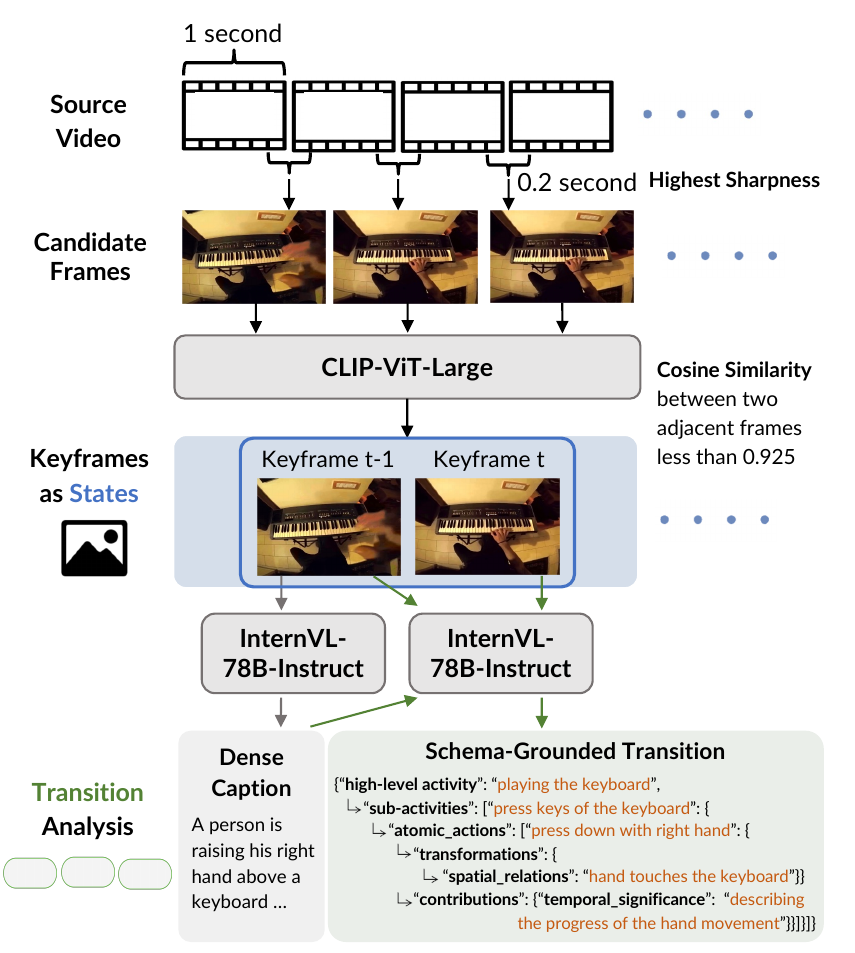}
\caption{\textbf{Overview of \vwm{} data construction.} We extract video keyframes as our visual states, ensuring high image sharpness and low CLIP embedding similarity across selected keyframes. Based on that, we prompt an oracle VLM (InternVL-78B-Instruct) to analyze the transitions between adjacent states.}
\label{data_construction}
\end{figure}

Figure~\ref{data_construction} presents an overview of how we construct the interleaved state-transition data for training our \vwm{} model. We describe the data construction pipeline below.

\textbf{State Extraction.} \quad
As videos naturally illustrate visual dynamics, we extract video keyframes to serve as the visual states in our training data.
To cover visual dynamics across broad domains, we select a diverse collection of source videos, including: (a) Ego4D \citep{grauman2022ego4d} videos that record ego-centric human daily activities in household, workplace, \textit{etc.}, (b) AgiBotWorld-Alpha \citep{bu2025agibot} videos that contain robot-view trajectories in environments such as restaurant, office, \textit{etc.}, (c) ShareGPT4Video \citep{chen2024sharegpt4video} that collects various real-world videos such as cooking tutorials, auto-driving recordings, and aesthetically appealing stock videos.\footnote{We exclude the Ego4D portion of videos collected in ShareGPT4Video to avoid duplication.}

We follow ShareGPT4Video \citep{chen2024sharegpt4video} to extract keyframes from the above videos in two steps.
First, we select candidate keyframes of each video by down-sampling
its frames to $\sim$$1$-second intervals.
To avoid selecting blurred candidates in the course of the video, our down-sampling interval is not fixed as exactly $1$ second, but allows a $\pm0.1$-second window of adjustment, as shown in Figure~\ref{data_construction}.
We traverse all frames within each $1\pm0.1$-second interval window and select the frame with the highest sharpness measured by Laplacian variance \citep{laparra2016perceptual}.
We then traverse the sequence of sampled candidates, and select the final keyframes by keeping only the candidate whose content is significantly different from its preceding selected keyframe.\footnote{The first candidate is always selected as a final keyframe.}
Specifically, as shown in Figure~\ref{data_construction}, we calculate the cosine similarity of CLIP \citep{radford2021learning} embeddings of the candidate and its preceding selected keyframe, and select the candidate as a keyframe if the similarity is lower than a threshold ($0.925$).

In Table~\ref{tab:state_data_stats_appdx} (Appendix~\ref{sec:data_cons_appdx}), we present detailed statistics of the source videos and extracted visual states (keyframes) that are included in the \vwm{} pre-training data.
The extracted visual state sequences cover diverse lengths (from $2$ to $100$ states per video) and content, due to the diversity of the original source videos.
Our state extraction is also monitored by the sharpness measure and CLIP embedding similarity, which ensures the image quality and allows flexible time intervals (from $0.8$ to $167.1$ seconds) between adjacent states.
The flexibility of time intervals enables \vwm{} to learn both \textbf{short-term} and \textbf{long-term} visual dynamics, such as \textit{a person's motion in a few seconds} and \textit{a long-term monitoring of a sunset}, respectively.


\begin{figure*}[t]
\centering
\includegraphics[width=1.0\textwidth]{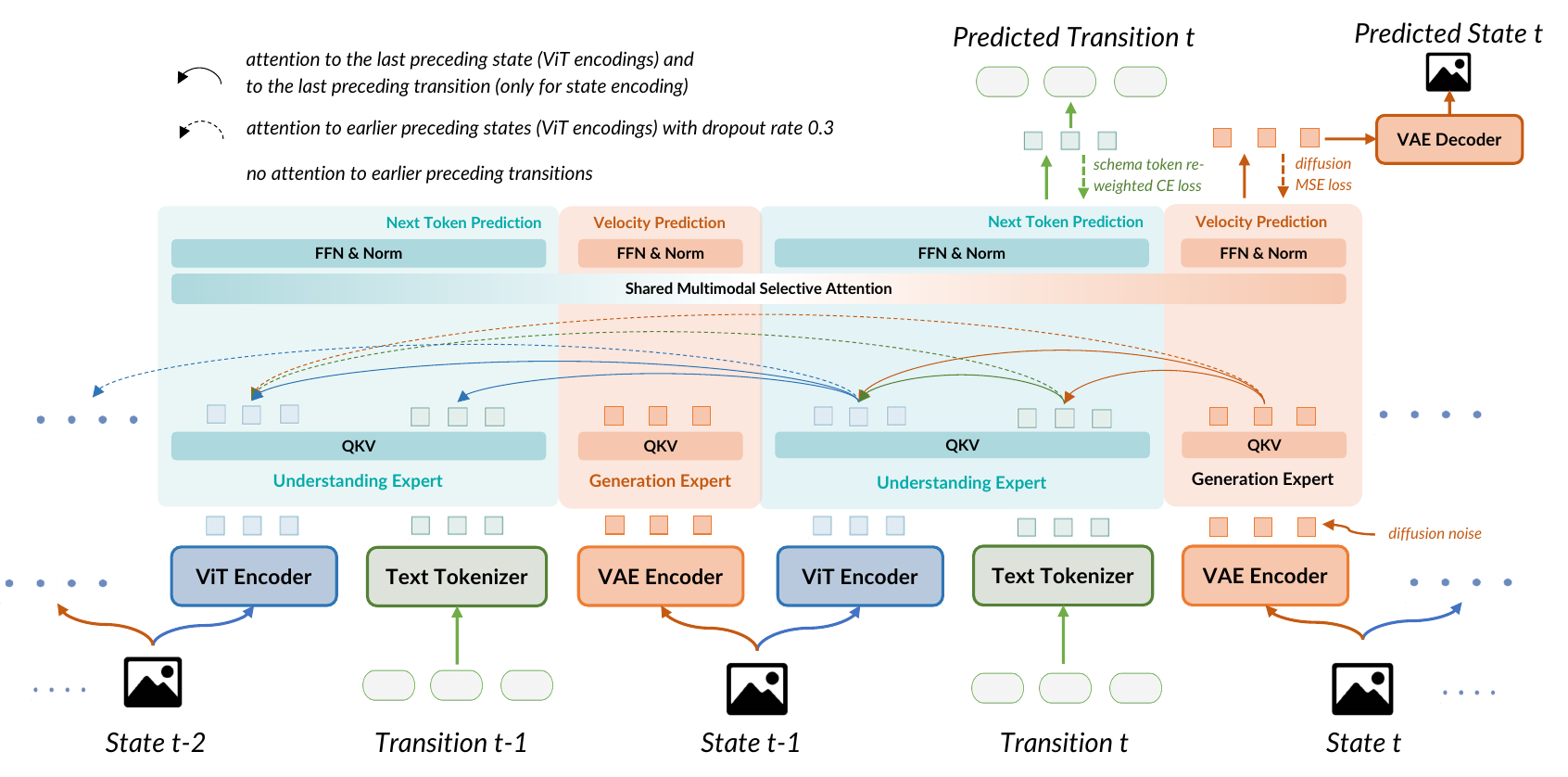}
\caption{\textbf{Overview of \vwm{} modeling approach.} The architecture of \vwm{} adopts a mixture of world understanding and generation experts, with shared multimodal selective attention. \vwm{} is jointly trained on the transition prediction and the state simulation, with a schema token re-weighted cross-entropy (CE) loss and a diffusion mean squared error (MSE) loss, respectively.}
\label{dynaview_modeling}
\end{figure*}

\textbf{Transition Frame and Annotation.} \quad
For each extracted sequence of visual states, we annotate the transition between each pair of adjacent states in the sequence.
To build hierarchical descriptions of visual dynamics, we frame the transition based on a fine-grained JSON schema, as illustrated in Figure~\ref{data_construction}.
Our transition schema captures both the progress of \textbf{high-level} activities (or events) and the changes of \textbf{low-level} visual details, with analysis of how the low-level changes contribute to the high-level progress.

\textbf{(a) High-Level Activity} describes the overarching scenario occurring between the adjacent states, \textit{i.e.,} a general description of the primary action or event taking place;
\textbf{(b) Sub-Activities} are constituents of the high-level activity, such as intermediate-level actions or events that contribute to the overall activity;
\textbf{(c) Atomic Actions} break down each sub-activity into the smallest meaningful units that cannot be further decomposed while maintaining semantic meaning.

Each atomic action can result in one or more of the following transformations when transitioning between the adjacent states:
\textbf{(d.1) Object Transformations} capture whether an object is newly introduced into, removed from or kept persistent in the scene;
\textbf{(d.2) Object State Transformations} detect whether the color, shape, texture, or physical state of any object is changed;
\textbf{(d.3) Spatial Relation Transformations} describe whether the positional relationship, contact or alignment of objects is altered;
\textbf{(d.4) Action Transformations} capture whether any continuation, completion, initiation or interruption of action has occurred;
\textbf{(d.5) Motion Transformations} detect whether any translational, rotational, oscillatory or deformation motion has happened;
\textbf{(d.6) Camera Transformations} describe whether the camera has any viewpoint changes, zoom operations, pan or tilt movements, or focus adjustments;
\textbf{(d.7) Background Transformations} detect whether the background has experienced any scene shifts, or changes of illumination, atmospheric conditions or contextual elements.

We analyze the overall contributions of above transformations from three aspects:
\textbf{(e.1) Atomic Action Contribution} describes how the transformations enable, facilitate, or result from the atomic action;
\textbf{(e.2) Sub-Activity Contribution} analyzes how the transformations support or advance the broader sub-activity goal;
\textbf{(e.3) Temporal Significance} captures the timing and sequence importance of the transformations within the overall high-level activity flow.\footnote{We present a complete example of our state transition schema in Figure~\ref{data_example} in Appendix~\ref{sec:data_cons_appdx}.}

Based on above schema, we prompt InternVL-78B-Instruct \citep{chen2024internvl} to annotate each state transition in two steps, as shown in Figure~\ref{data_construction}.
Given a pair of adjacent states, we first prompt InternVL to generate a dense caption of the former state, which guides the model to understand the events, actions and visual details shown at the start of the state transition.
On top of the dense caption, we then instruct InternVL to compare the latter state with the former state, and generate transition descriptions following our defined schema.
To ensure that the output of InternVL is in valid JSON structure, we use constrained decoding \citep{hokamp2017lexically} to restrict the token generation.\footnote{We include our JSON schema used for constrained decoding and prompts for transition annotation in Appendix~\ref{sec:data_cons_appdx}.}

Our hierarchical transition descriptions enable \vwm{} to learn a comprehensive understanding of how human world events or actions and their associated visual elements are coherently arranged and transformed over time.
Moreover, the fine-grained verbalization of transitions also facilitates \vwm{}'s intermediate thinking and planning of the latter visual state simulation, which contributes to better controllability of visual generation.
These promote \vwm{} to build a more general and reliable world modeling.

\subsection{\vwm{} Modeling Approach}
\label{sec_model}
Figure~\ref{dynaview_modeling} shows the overall architecture of \vwm{}. 
We adopt a mixture-of-Transformer-experts (MoT) architecture following BAGEL \citep{deng2025emerging}, to integrate both world understanding and generation capabilities into \vwm{}.
For world understanding, each state that provides contexts for predicting subsequent transitions and states is encoded by a vision Transformer (ViT), using the SigLIP2 (so400m-patch14-384; \citealp{tschannen2025siglip}) architecture with the NaViT \citep{dehghani2023patch} approach of preserving the native image aspect ratio.
The ViT encodings of state tokens (or patches) and the embeddings of transition tokens are both routed to an \textit{understanding} expert with Qwen2.5 (7B; \citealp{qwen2025qwen25tech}) LLM  architecture.
For world generation, a variational autoencoder (VAE) FLUX (1-dev; \citealp{flux2024,labs2025flux}) is used to encode and decode each state that needs to be predicted.
The VAE encodings of state tokens, after corruption with diffusion noise for rectified-flow training\footnote{To save the token budget of \vwm{} and thus enable longer state-transition sequence modeling, we exclude the noise-free VAE encodings used as additional decoding conditions in BAGEL, which have minor impact on improving \vwm{} learning.}, are routed to a \textit{generation} expert with the same Qwen2.5-7B architecture.
The understanding and generation experts perform shared multimodal self-attention at every Transformer layer.

\textbf{Multimodal Selective Attention.} \quad
We employ a selective mask in the multimodal self-attention of \vwm{}, as shown in Figure~\ref{dynaview_modeling}, to reduce redundant information and avoid naive failure modes in state-transition sequence modeling.
Specifically, we mask out each transition's attention to its preceding transitions, and only keep the attention to preceding states' ViT encodings that implicitly reveal the historical transitions.
This mask prevents \vwm{} from naively copying one of the historical transitions as the prediction.
Similarly, we limit each state's attention (either ViT- or VAE-encoded) to (a) the last transition that leads to the state, and (b) its preceding states' ViT encodings. We mask out the attention to earlier transitions to avoid interference with the state simulation.
To mitigate potential overfitting to our training data distribution, we also employ a dropout of attention to the preceding states.
In particular, for each state (or transition), we do not drop out its attention to the last state preceding it, which contains the most recent visual contexts for prediction, but we mask out the attention to each of the earlier states with a probability of $0.3$.
Finally, we adopt causal attention within each transition's text tokens, and bidirectional attention within each state's ViT or VAE tokens.

\begin{table*}[t]
\centering
\caption{Validation of \vwm{} transition prediction and state simulation, compared to the gold references, using GPT-4o as the judge.
}
\resizebox{1.0\textwidth}{!}{
\smallskip\begin{tabular}{lccccccccccc}
\toprule
 \multirow{3}*{\textbf{Method}} & \multicolumn{7}{c}{\textbf{Transition Prediction}} & \multicolumn{4}{c}{\textbf{State Simulation}} \\
\cmidrule(lr){2-8}  \cmidrule(lr){9-12}
 & \multicolumn{2}{c}{\textbf{Activities}} & \multicolumn{3}{c}{\textbf{Transformations}} & \multicolumn{2}{c}{\textbf{Contributions}} & \textbf{Style Consistency} & \multicolumn{3}{c}{\textbf{Depicted Transformations}} \\
 \cmidrule(lr){2-3} \cmidrule(lr){4-6} \cmidrule(lr){7-8} \cmidrule(lr){9-9} \cmidrule(lr){10-12}
 & Accept (\%) & Reject (\%) & Accept (\%) & Reject (\%) & None (\%) & Accept (\%) & Reject (\%) & Score (1-10) & Accept (\%) & Reject (\%) & None (\%) \\
\toprule
 \vwm{} & 98.11 & 1.89 & 43.57 & 0.53 & 55.90 & 98.00 & 2.00 & 8.54 & 65.39 & 3.48 & 31.13 \\
 Gold & 98.89 & 1.11 & 44.19 & 0.43 & 55.38 & 98.67 & 1.33 & 8.89 & 58.89 & 1.22 & 39.89 \\
\bottomrule
\end{tabular}
}
\label{tab:vwm_validation}
\end{table*}

\textbf{Training Objectives.} \quad
We initialize \vwm{} with the pre-trained weights of BAGEL-7B-MoT \citep{deng2025emerging}, and continue pretraining on both transition prediction and state simulation tasks.
The two tasks are performed alternately in an interleaved sequence, as shown in Figure~\ref{dynaview_modeling}.
\vwm{} performs next token prediction on top of the understanding expert to auto-regressively generate each transition, which is trained on the cross-entropy (CE) loss.
Note that our defined transition consists of JSON-schema tokens with only slight variations across different training samples, and slot-filling tokens that can vary dramatically across different samples, as illustrated by the \textbf{black} and \textcolor{myorange}{orange} tokens in Figure~\ref{data_construction}, respectively.
To prevent \vwm{} from overfitting to the relatively static (and thus more frequently appeared) JSON-schema tokens, while under-learning the more dynamically varied slot-filling tokens, we adopt a \textbf{schema token re-weighted CE loss} for training the transition token prediction.
Specifically, we parse each gold reference transition to identify its JSON-schema tokens, and set a lower CE loss weight ($0.1$) on these identified tokens, while keep a normal CE loss weight ($1.0$) on the rest of slot-filling tokens.
This balances \vwm{}'s learning of the transition structure and the more specific transition details.
For state simulation, \vwm{} performs velocity prediction \citep{salimans2022progressive} on top of the generation expert, to simulate the de-noised VAE encodings of each state image.
The predicted de-noised encodings are then compared to the gold state's clean VAE encodings, to calculate the mean squared error (MSE) loss for diffusion training.
At the inference phase, pure Gaussian noise is sampled as the input to the generation expert for each state simulation, and the VAE decoder is used to generate the simulated state image based on the predicted de-noised encodings.
We sum the CE loss for transition prediction and the MSE loss for state simulation with equal weights ($1.0$) as \vwm{}'s final training loss.\footnote{More details of \vwm{}'s training and inference methods are presented in Appendix~\ref{sec:model_appdx}.}

\subsection{\vwm{} Validation}
\label{sec_valid}
In this section, we validate \vwm{}'s performance of world modeling and the quality of our state-transition data construction.
Specifically, we sample $900$ more source videos, $300$ each from Ego4D, AgiBotWorld-Alpha and ShareGPT4Video datasets (excluding videos in the training data), and construct their state-transition data using the pipeline described in Section~\ref{sec_data}.
Based on that, we use GPT-4o \citep{hurst2024gpt,achiam2023gpt} as a judge to evaluate \vwm{}'s predicted transitions and simulated states on these $900$ validation samples, and to check the gold references constructed by our pipeline.

For each predicted (or gold) \textit{transition}, we prompt the judge to check the filled values in each transition slot defined in our JSON schema, including three levels of \textbf{Activities} (high-level activity, sub-activities and atomic actions), seven types of \textbf{Transformations} (objects, object states, spatial relations, actions, motion, camera and background), and three types of \textbf{Contributions} (to atomic action, to sub-activity and temporal significance).
The judge evaluates whether (Accept) or not (Reject) each slot value describes a reasonable prediction given the transition's preceding states as contexts.
A ``None'' result is given if a slot is empty with no annotation.

For each simulated (or gold) \textit{sequence of states}, we instruct the judge to first score the \textbf{Style Consistency} of the states on a Likert scale from 1 to 10 (higher is better).
Similar to the evaluation of Transformations, the judge then evaluates each state in the sequence, \textit{w.r.t.} whether (Accept) or not (Reject) each of the state's \textbf{Depicted Transformations} relative to its preceding states are reasonable.
A ``None'' result is given if a type of Transformation is not depicted in the state.\footnote{We include our judge's validation prompts in Appendix~\ref{sec:valid_appdx}.}

Table~\ref{tab:vwm_validation} shows our validation results on \vwm{} outputs and gold references.
\vwm{} predicts transitions with fairly low Reject rates on all three aspects, which are not exceeding $2$\% and comparable to the gold annotations, indicating that the transitions predicted by \vwm{} are reliable and the annotated gold transitions are also accurate.
For state simulation, \vwm{} also achieves a high style consistency score (above $8.5$ out of $10$, comparable to gold states) and a low Reject rate (less than $5$\%) of depicted transformations, indicating a decent capability for visual dynamic sequence generation.
Although the precision of \vwm{}'s simulated transformations slightly falls behind gold references, with $\sim2$\% higher Reject rate, it generates richer transformations as a trade-off for higher recall, with higher Accept rate and lower None rate.
Our validation results verify the effective learning of \vwm{} based on our high-quality state-transition data.

\section{Experimental Methods}

Robustly learning visual dynamics boosts \vwm{}'s ability to be adapted for various downstream multimodal tasks.
We test the advantages of \vwm{} on several downstream benchmarks that focus on two complementary capabilities: visual narrative generation \citep{gao2025vinabench} and instruction following in world simulation \citep{lai2024lego}.

\textbf{Visual Narrative Generation.} \quad
Our primary evaluation targets the visual narrative generation task, \textit{i.e.}, creating
continuations of an image sequence to depict the scenes of a narrative, which requires understanding narrative-driven state transitions, long-horizon consistency across narrative images, and progression of causal events.
We adopt the Visual Writing Prompts (VWP) portion of VinaBench~\citep{gao2025vinabench} as our evaluation benchmark, which covers diverse scenes from movies.
Each VinaBench sample consists of a sequence of images depicting a visual narrative's scenes, where each image is aligned with a textual description and a constraint capturing key elements (\textit{e.g.}, time of day, characters, location) of the image, as well as connections between visual and textual entities (such as linking characters in the image to their names mentioned in the text).

Given an initial narrative image and a textual description of the next scene, the model must generate a realistic and coherent continuation image to depict the next scene.
The generated image and its description are appended to the context, enabling iterative prediction along the full narrative trajectory.
We test model performance under both zero-shot generalization and supervised fine-tuning (SFT) settings.
In the SFT setting, the model is fine-tuned on VinaBench training samples to also predict the constraint of the next scene before simulating the next scene image.
Based on the constraint of each scene, we follow \citet{gao2025vinabench} and evaluate the model on two per-scene text-image alignment and cross-scene consistency.
We employ oracle LLMs, GPT-4o \citep{hurst2024gpt} and Gemini-2.5-Pro \citep{comanici2025gemini}, as the judges to measure alignment and consistency, based on VQAScore \citep{lin2024evaluating} and judgment questions \textit{w.r.t.} the constraint.\footnote{We introduce more details of the baseline models, benchmarks and VQA-based evaluation in Appendix~\ref{sec:baseline_methods},~\ref{sec:eval_benchmarks}~and~\ref{sec:vqa_prompts}.}

\textbf{Instruction Following in World Simulation.} \quad
In contrast to long-horizon and open-ended visual narrative generation, where multiple scene continuations may be valid, world simulation requires translating textual instruction (or action) into concrete state changes, leading to largely deterministic outcomes.
We evaluate this capability on LEGO \citep{lai2024lego}, where the world model predicts the next visual state based on an image representation of the current state and a textual prompt specifying the action.
We follow LEGO and use three evaluation metrics based on contrastive learning: CLIP \citep{radford2021learning} score, EgoVLP and EgoVLP$^+$ \citep{lin2022egocentric}, and three common image generation metrics, FID \citep{heusel2017gans}, PSNR and LPIPS \citep{zhang2018unreasonable} (with SqueezeNet proposed by \citealp{iandola2016squeezenet} as the encoder), to make a thorough evaluation.

% ...

\begin{table*}[t]
\centering
\caption{
Evaluation results of visual narrative generation on VinaBench, with GPT-4o as the judge.
``Non-Char. Ent.'', ``Char. Num.'' and ``Char. Attr.'' indicate Non-Character Entities, Character Number and Character Attribute, respectively.
The best results within each block are \textbf{in bold} and the second-best are \underline{underlined}. 
More details of different inference settings are included in Appendix~\ref{sec:inference_settings}.
}
\resizebox{1.0\textwidth}{!}{
\smallskip
\begin{tabular}{lccccccccc}
\toprule
\multirow{2}*{\textbf{Model}}
& \multicolumn{5}{c}{\textbf{Per-Scene Alignment}} 
& \multicolumn{3}{c}{\textbf{Cross-Scene Consistency}} 
& \multirow{2}*{\textbf{Average}}
\\
\cmidrule(lr){2-6} \cmidrule(lr){7-9}
& Non-Char. Ent. & Char. Num. & Char. Attr. & Time & Location
& Style & Character & Location & 
\\
\midrule
\multicolumn{10}{c}{\textbf{Zero-Shot Generalization}} \\
\midrule
Emu2
& 0.533 & 0.220 & \underline{0.556} & 0.523 & 0.466 
& 0.378 & 0.352 & 0.439
& 0.425 \\

BAGEL
& \underline{0.622} & 0.315 & 0.528 & \textbf{0.599} & \underline{0.492} 
& 0.824 & 0.370 & \underline{0.675} 
& \underline{0.567} \\

Story2Board
& \textbf{0.683} & \textbf{0.353} & \textbf{0.611} & 0.392 & 0.374 
& \textbf{0.984} & \underline{0.472} & 0.490 
& 0.566 \\

\rowcolor{gray!20}
DynaVieW
& 0.600 & \underline{0.323} & 0.521 & \underline{0.551} & \textbf{0.501} 
& \underline{0.897} & \textbf{0.550} & \textbf{0.835} 
& \textbf{0.630} \\

\midrule
\multicolumn{10}{c}{\textbf{Supervised Fine-Tuning}} \\
\midrule
ARLDM 
& 0.701 & 0.403 & 0.621 & 0.462 & 0.461
& \underline{0.806} & 0.313 & 0.329
& 0.506 \\

MM-Interleaved
& 0.659 & \textbf{0.425} & \underline{0.625} & 0.518 & 0.493
& 0.795 & 0.336 & 0.404
& 0.528 \\

StoryGen
& 0.566 & 0.388 & 0.412 & 0.313 & 0.416
& 0.576 & 0.158 & 0.341
& 0.389 \\

BAGEL
& \underline{0.723} & \underline{0.420} & 0.622 & \underline{0.533} & \textbf{0.512}
& 0.770 & \underline{0.344} & \underline{0.483}
& \underline{0.547} \\

\rowcolor{gray!20}
DynaVieW
& \textbf{0.726} & 0.414 & \textbf{0.631} & \textbf{0.549} & \underline{0.511}
& \textbf{0.879} & \textbf{0.483} & \textbf{0.601}
& \textbf{0.610} \\

\bottomrule
\end{tabular}
}
\label{tab:vinabench_score_main}
\end{table*}

\begin{table*}[t]
\centering
\caption{
Evaluation results of controllability on visual narrative generation, with GPT-4o as the judge.
We use Gemma-3 (27B-it) as an assistant to generate a JSON-schema description of the next scene for multiple aspects, enabling fine-grained control over next-scene image simulation in visual narrative models.
Since BAGEL is not pre-trained on JSON data, we further transform each JSON-schema description into natural language (NL) and use these to prompt visual generation.
We report standard deviation across three random seeds.
The best results within each block are \textbf{in bold} and the second-best results are \underline{underlined}. 
More details of generated transitions are included in Appendix~\ref{sec:inference_settings}.
}
\resizebox{1.0\textwidth}{!}{
\smallskip
\begin{tabular}{@{~}lc@{~~~}c@{~~~}c@{~~~}c@{~~~}cc@{~~~}c@{~~~}cc@{~}}
\toprule
\multirow{2}*{\textbf{Model}}
& \multicolumn{5}{c}{\textbf{Per-Scene Alignment}} 
& \multicolumn{3}{c}{\textbf{Cross-Scene Consistency}} 
& \multirow{2}*{\textbf{Average}}
\\
\cmidrule(lr){2-6} \cmidrule(lr){7-9}
& Non-Char. Ent. & Char. Num. & Char. Attr. & Time & Location
& Style & Char. & Location & 
\\
\midrule
BAGEL-NL
& \meanstd{0.355}{0.005} & \meanstd{0.283}{0.001} & \meanstd{0.301}{0.003} & \meanstd{0.433}{0.015} & \meanstd{0.305}{0.002} 
& \meanstd{0.327}{0.010} & \meanstd{0.108}{0.003} & \meanstd{0.207}{0.006} 
& \meanstd{0.275}{0.001} \\

BAGEL-Schema
& \meanstd{0.355}{0.006} & \meanstd{0.288}{0.008} & \meanstd{0.306}{0.007} & \meanstd{0.485}{0.003} & \meanstd{0.307}{0.003} 
& \meanstd{0.351}{0.012} & \meanstd{0.109}{0.002} & \meanstd{0.194}{0.004} 
& \meanstd{0.283}{0.002} \\

\rowcolor{gray!20}
\vwm{}-NL
& \meanstd{\underline{0.530}}{0.008} & \meanstd{\textbf{0.325}}{0.010} & \meanstd{\underline{0.502}}{0.004} & \meanstd{\underline{0.493}}{0.011} & \meanstd{\textbf{0.476}}{0.002} 
& \meanstd{\underline{0.803}}{0.012} & \meanstd{\textbf{0.588}}{0.007} & \meanstd{\textbf{0.796}}{0.007}
& \meanstd{\underline{0.597}}{0.002} \\

\rowcolor{gray!20}
\vwm{}-Schema
& \meanstd{\textbf{0.541}}{0.006} & \meanstd{\underline{0.317}}{0.004} & \meanstd{\textbf{0.509}}{0.009} & \meanstd{\textbf{0.510}}{0.011} & \meanstd{\underline{0.471}}{0.004}
& \meanstd{\textbf{0.858}}{0.013} & \meanstd{\underline{0.564}}{0.001} & \meanstd{\underline{0.792}}{0.013}
& \meanstd{\textbf{0.604}}{0.004} \\

\bottomrule
\end{tabular}
}
\label{tab:vinabench_score_controllability}
\end{table*}

\section{Experimental Results}

\textbf{\vwm{} Excels at Visual Narrative Generation.} \quad
In Table~\ref {tab:vinabench_score_main}, we present our evaluation results on VinaBench using GPT-4o as the judge.\footnote{Results with Gemini-2.5-Pro as the judge are presented in Table~\ref{tab:vinabench_score_gemini} in Appendix~\ref{sec:gemini_2.5_pro_eval_results}, which draw the same conclusions.}
In the zero-shot generalization setting, \vwm{} achieves higher Average score than Emu2~\citep{sun2024generative}, BAGEL~\citep{deng2025emerging} and a strong baseline Story2Board \citep{dinkevich2025story2board} optimized for storyboard creation, driven by clear advantages in cross-scene consistency, while remaining competitive on most per-scene alignment metrics.
\vwm{} achieves strong location consistency, which we attribute to extensive pre-training on data that emphasizes visual dynamics at fixed spatial positions.

With access to domain-specific data, \vwm{} (SFT) attains the highest Average score (and also the best or second-best stratified scores) across all SFT baselines, including ARLDM \citep{pan2024synthesizing}, MM-Interleaved \citep{tian2024mm}, StoryGen \citep{liu2024intelligent} and BAGEL.
Compared to strong baselines such as BAGEL, \vwm{} shows a more balanced improvement across both per-scene fidelity and long-range consistency, indicating that its schema-guided state-transition modeling effectively structures generation and reduces error accumulation. 
Compared to zero-shot performances, SFT trades off cross-scene consistency scores with higher per-scene alignment scores, indicating richer visual manifestations related to the narrative, instead of coherent but less expressive visual narrative creation.
Overall, these results suggest that explicitly modeling the next visual state before generation is particularly beneficial for complex visual narratives that require both fine-grained accuracy and global coherence.

We conduct a head-to-head human evaluation between \vwm{} (SFT) and BAGEL (SFT) to further validate our experiment results. 
We ask 8 human testers to read the input textual narratives, and then compare storyboards generated by the two models (anonymously presented and randomly ordered). 
In total, we collected 400 preference votes. The results show that human testers prefer \vwm{} in 47.6\% of cases, compared to 35.4\% for BAGEL, with 17\% of comparisons resulting in a tie. The human evaluation results align well with automatic VLM judges. 

\textbf{Controllability from Hierarchical Modeling.} \quad
Beyond visual generation performance, we further evaluate the controllability of visual generation. We employ Gemma-3 (27B-it; \citealp{team2025gemma}),
to predict a JSON-schema description of transitions to the next narrative image (or state), exemplified in Figure~\ref{data_construction}.
By sourcing transitions externally rather than from the \vwm{} model, we decouple the rendering of a given transition from the model’s own transition prediction process, thereby providing a more accurate measurement of the model’s transition steerability. 
Moreover, diverse and even slightly noisy transition specifications obtained from Gemma-3 (27B-it) allows us to test whether DynaVieW can robustly simulate states under a broad range of possible transition conditions, including partially infeasible ones.
In Table~\ref{tab:vinabench_score_controllability}, the evaluation results show that \vwm{} enhanced by LLM-generated JSON schemas substantially outperforms BAGEL across all metrics (0.604 vs 0.283 on average).
Note that transforming JSON schemas into natural language (via GPT-4o) does not improve BAGEL’s overall performance.
Its average score remains significantly lower than those of \vwm{} (0.597 vs 0.275), underscoring that the benefit arises from the hierarchical visual dynamic modeling rather than the representation format. 
These results demonstrate that hierarchical world modeling provides a more controllable generation process, enabling models to better translate high-level constraints into coherent and faithful visual narratives.

\textbf{Strong Visual Instruction Following.} \quad
In Table~\ref{tab:lego_instruction_following}, we summarize the instruction following performances of \vwm{} and BAGEL on the LEGO benchmark, which contains two testing subsets (Ego4D and Epic-Kitchens).
Models are evaluated both before and after SFT on LEGO's task-specific training data.
On Ego4D, \vwm{} consistently achieves better performance compared to BAGEL across all evaluation metrics, indicating improvements in both visual fidelity and semantic alignment with instructions.
Notably, these gains persist and become more pronounced after SFT. 
For example, \vwm{} (SFT) attains an FID of 20.96, corresponding to a 19.8\% reduction compared to \vwm{} and a 4.8\% improvement over BAGEL (SFT), highlighting its ability to follow fine-grained, action-centric instructions in complex egocentric scenarios. 
The advantage of \vwm{} in world simulation is also evident on Epic-Kitchens, beating BAGEL in nearly all metrics.
In particular, \vwm{} surpasses BAGEL on CLIP (77.79 vs 76.76), with the performance gap becoming larger after SFT.
Moreover, we validate the superiority of our model on two image-to-text metrics with results reported in Appendix~\ref{sec:lego_image_to_text_score} (Table~\ref{tab:lego_image_to_text_score}).
It indicates that \vwm{} provides a stronger foundation for downstream world simulation tasks.
Overall, these results suggest that modeling visual dynamics explicitly enables \vwm{} to better interpret and execute visual instructions, leading to more faithful, semantically aligned, and perceptually coherent outputs. 
We provide additional comparisons on Epic-Kitchens against baselines from prior work in Appendix~\ref{sec: more_baselines}.
We also present several model output examples in Figure~\ref{fig:storyboard_creation_examples} and Figure~\ref{fig:lego_examples} (Appendix~\ref{sec:benchmark_examples}).

% ...

\begin{table}[t]
\centering
\caption{Results of instruction following in world simulation on LEGO.
The best results within each testing subset are \textbf{in bold}. 
}
\resizebox{0.5\textwidth}{!}{
\smallskip
\begin{tabular}{@{~}lc@{~~~}c@{~~~}c@{~~~}c@{~~~}c@{~~~}c@{~}}
\toprule
& FID $\downarrow$
& PSNR $\uparrow$
& LPIPS $\downarrow$
& CLIP $\uparrow$
& EgoVLP $\uparrow$
& EgoVLP$^+$ $\uparrow$ \\
\midrule
\multicolumn{7}{c}{\textbf{Ego4D}} \\
\midrule
BAGEL
& 26.94 & 10.64 & 43.57 & 73.78 & 45.07 & 72.61 \\

\rowcolor{gray!20}
\vwm{}
& 26.15 & 10.74 & 43.47 & 76.09 & 52.88 & 74.54 \\

BAGEL (SFT)
& 22.02 & 11.18 & 41.57 & 79.37 & 55.32 & 76.16 \\

\rowcolor{gray!20}
\vwm{} (SFT)
& \textbf{20.96} & \textbf{11.43} & \textbf{40.77} & \textbf{80.99} & \textbf{58.73} & \textbf{77.62} \\

\midrule
\multicolumn{7}{c}{\textbf{Epic-Kitchens}} \\
\midrule

BAGEL
& 20.75 & 10.88 & 41.45 & 76.76 & 40.65 & 58.41 \\

\rowcolor{gray!20}
\vwm{}
& 18.33 & \textbf{11.48} & 41.78 & 77.79 & 43.60 & 58.83 \\

BAGEL (SFT)
& 11.66 & 11.11 & 39.91 & 82.71 & 47.46 & 61.21 \\

\rowcolor{gray!20}
\vwm{} (SFT)
& \textbf{10.31} & 11.31 & \textbf{39.20} & \textbf{84.19} & \textbf{48.72} & \textbf{61.77} \\

\bottomrule
\end{tabular}
}
\label{tab:lego_instruction_following}
\end{table}

\begin{figure*}[t]
  \centering
  \includegraphics[width=\textwidth]{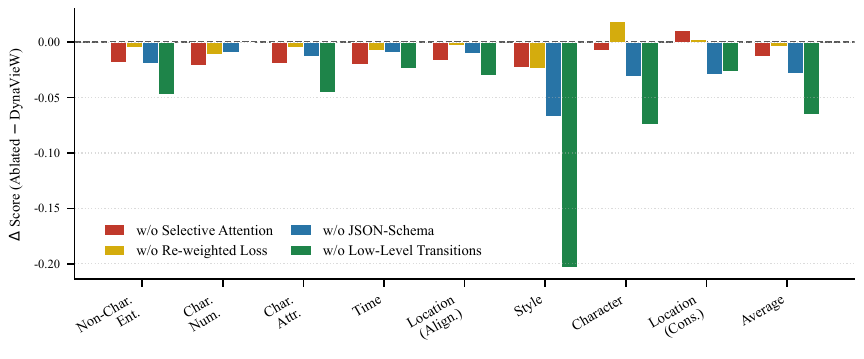}
  \caption{
    \textbf{Ablation study of each major component in \vwm{} on VinaBench.}
    We (a) remove the multimodal selective attention (w/o \textit{Selective Attention}), 
    (b) remove the schema token re-weighting of CE loss (w/o \textit{Re-weighted Loss}), 
    (c) use a heuristic template to translate the JSON-schema transitions into natural language descriptions (w/o \textit{JSON-Schema}),
    and (d) use a more coarse-grained transition schema without describing the fine-grained low-level visual transformations and their contributions (w/o \textit{Low-level Transitions}).
    Performance change ($\Delta$ score) when removing each component from DynaVieW.
    Negative values indicate performance drops relative to the full approach.
  }
  \label{fig:ablation}
\end{figure*}

\textbf{Ablation Study.} \quad
We conduct a detailed ablation study on the major components of our method:
(a) removing multimodal selective attention (w/o \textit{Selective Attention});
(b) removing schema-token re-weighting in the CE loss (w/o \textit{Re-weighted Loss});
(c) replacing the JSON-schema transitions in the training data with natural-language descriptions generated from a heuristic template (w/o \textit{JSON-Schema}); and
(d) using a coarser-grained transition schema that omits fine-grained low-level visual transformations and their contributions during training (w/o \textit{Low-level Transitions}).
We compare the original DynaVieW with its ablated variants on the downstream VinaBench dataset under the zero-shot generalization setting used in Table~\ref{tab:vinabench_score_main}.
As shown in Figure~\ref{fig:ablation}, removing multimodal selective attention consistently degrades image-text alignment, while its impact on cross-scene consistency is mixed.
Removing schema-token re-weighting in the CE loss also underperforms the full DynaVieW model, albeit with a smaller average drop than removing multimodal selective attention (-0.004 vs. -0.013).
Our ablation study on the training data further highlights the importance of both JSON-structured descriptions of visual dynamics and fine-grained hierarchical transitions for effective DynaVieW training. Ablating these components yields average decreases of 0.028 and 0.064, respectively.
In particular, using only coarse-grained JSON transitions substantially degrades the model's ability to preserve cross-scene style and character consistency, revealing the central role of fine-grained transition granularity.

\section{Related Work}

\textbf{World Modeling} \quad
World models are typically tasked with understanding the present state of the world and predicting its future dynamics \citep{ding2025understanding}.
Early world models were proposed for simulating spatial-temporal representations of game environments, to facilitate reinforcement learning of game agents \citep{ha2018world}.
Subsequently, the scope of world models was extended to real-world environment simulation \citep{yang2023learning}, \textit{i.e.}, and predicting possible future real-world states as a function of imagined actions \citep{lecun2022path}.
More recent studies have shown that large language models (LLMs) possess promising world modeling abilities \citep{gurnee2024language}, which is closely related to their capabilities for reasoning and planning \citep{hao2023reasoning}, and visual generation \citep{liu2024sora}.
LLMs pre-trained with world modeling objectives have been applied to various domains, such as robot learning \citep{wu2023daydreamer}, social interaction simulacra \citep{park2023generative} and auto-driving \citep{gao2024vista}.
Our work builds on these foundations to represent more complex visual dynamics in our world model training objective, resulting in improved reliability and controllability.

\textbf{Visual Dynamics} \quad
The study of visual dynamics is crucial for understanding human activities, from dynamics scenes of daily life \citep{pittore2000learning,buxton2003learning} to visual narratives \citep{nienhaus2005depicting}, and also provides an essential ground for image or video frame sequence synthesis and prediction \citep{xue2016visual,finn2016unsupervised}.
However, interpreting world visual dynamics requires multiple aspects of visual cognition, including spatial-temporal grouping of objects \cite{gepshtein2000emergence}, tracking of object state changes \citep{dickmanns2007dynamic}, the perception of motion and actions \citep{johansson1975visual,bonnet2005visual,dickmanns2007dynamic}, and modeling background and camera control \citep{rowe1995statistical,yao2025beyond}.
We integrate all above aspects of visual dynamic cognition into our hierarchical frame of transition descriptions, to promote more comprehensive understanding of visual dynamics.

\textbf{Interleaved Vision-Text LLMs} \quad
Interleaved vision-text data was originally used for training multimodal LLMs capable of few-shot learning \citep{alayrac2022flamingo}, and is now a typical datastream for unifying visual understanding and generation \citep{tian2024mm}.
Early interleaved vision-text LLMs \citep{tian2024mm,team2024chameleon,chern2024anole,sun2024generative,dong2024dreamllm} were pre-trained on public web-sourced image-text sequences \citep{zhu2023multimodal,laurenccon2023obelics,sun2024emu}.
However, the image sequences in web crawls often lacked alignment between adjacent images, \textit{e.g.}, illustrations of unrelated key entities in a paragraph, leading to poor understanding of the visual dynamics required by world modeling.
More advanced interleaved vision-text LLMs leverage frames from videos \citep{soldan2022mad,fan2022minedojo,bu2025agibot,wang2025koala} to construct the visual sequences.
Still, their source video types and textual annotations are mostly limited to a single aspect of visual dynamic cognition, such as high-level progression of movie plots \citep{lin2024learning}, viewpoint changes in game environments \citep{zhang2025matrix,he2025matrix}, actions of robots \citep{qu2025eo} and coarse-grained captions of scene shifts \citep{deng2025emerging}.
In this work, we build broader and more comprehensively annotated video data to elevate the world modeling capability of interleaved vision-text LLMs.

\section{Conclusion}
In this paper, we propose a new world model, \vwm{} to understand and generated complex visual dynamics. 
\vwm{} builds in-depth comprehension of dynamic visual scenes by joint learning to simulate video states and predict hierarchical attributes of state transitions. 
In downstream experiments, \vwm{} demonstrates more controllable visual generation and better instruction following compared to competitive baselines.
Our results highlight how training on world modeling objectives can improve multimodal LLMs, revealing a promising future direction for developing more powerful world models that can serve as reliable backbones of multimodal applications.





\section*{Acknowledgements}

We gratefully acknowledge the support of the Swiss National Science Foundation (No. 215390), the European Research Council (Starting grant no. 101222478, RESPECT-LM), the AI2050 program at Schmidt Sciences (Grant \#G-25-69783), Sony Group Corporation, and the Swiss National Supercomputing Center (CSCS) in the form of an infrastructure engineering and development project.
This research is with support from Google.org and the Google Cloud Research Credits program for the Gemini Academic Program.
This work was supported as part of the Swiss AI Initiative by a grant from the Swiss National Supercomputing Centre (CSCS) under project ID a173 on Alps.
Li Mi gratefully acknowledges the support of the Horizon Europe grant 101213369 DVPS.



\section*{Impact Statement}


This paper presents work whose goal is to advance the field of Multimodal Large Language Models in Artificial Intelligence.
There are many potential societal consequences of our work, none of which we feel must be specifically highlighted here.



\bibliography{main}

@inproceedings{grauman2022ego4d,
  title={Ego4d: Around the world in 3,000 hours of egocentric video},
  author={Grauman, Kristen and Westbury, Andrew and Byrne, Eugene and Chavis, Zachary and Furnari, Antonino and Girdhar, Rohit and Hamburger, Jackson and Jiang, Hao and Liu, Miao and Liu, Xingyu and others},
  booktitle={Proceedings of the IEEE/CVF conference on computer vision and pattern recognition},
  pages={18995--19012},
  year={2022}
}

@article{bu2025agibot,
  title={Agibot world colosseo: A large-scale manipulation platform for scalable and intelligent embodied systems},
  author={Bu, Qingwen and Cai, Jisong and Chen, Li and Cui, Xiuqi and Ding, Yan and Feng, Siyuan and Gao, Shenyuan and He, Xindong and Hu, Xuan and Huang, Xu and others},
  journal={arXiv preprint arXiv:2503.06669},
  year={2025}
}

@article{chen2024sharegpt4video,
  title={Sharegpt4video: Improving video understanding and generation with better captions},
  author={Chen, Lin and Wei, Xilin and Li, Jinsong and Dong, Xiaoyi and Zhang, Pan and Zang, Yuhang and Chen, Zehui and Duan, Haodong and Tang, Zhenyu and Yuan, Li and others},
  journal={Advances in Neural Information Processing Systems},
  volume={37},
  pages={19472--19495},
  year={2024}
}

@article{deng2025emerging,
  title={Emerging properties in unified multimodal pretraining},
  author={Deng, Chaorui and Zhu, Deyao and Li, Kunchang and Gou, Chenhui and Li, Feng and Wang, Zeyu and Zhong, Shu and Yu, Weihao and Nie, Xiaonan and Song, Ziang and others},
  journal={arXiv preprint arXiv:2505.14683},
  year={2025}
}

@article{laparra2016perceptual,
  title={Perceptual image quality assessment using a normalized Laplacian pyramid},
  author={Laparra, Valero and Ball{\'e}, Johannes and Berardino, Alexander and Simoncelli, Eero P},
  journal={Electronic Imaging},
  volume={28},
  pages={1--6},
  year={2016},
  publisher={Society for Imaging Science and Technology}
}

@inproceedings{radford2021learning,
  title={Learning transferable visual models from natural language supervision},
  author={Radford, Alec and Kim, Jong Wook and Hallacy, Chris and Ramesh, Aditya and Goh, Gabriel and Agarwal, Sandhini and Sastry, Girish and Askell, Amanda and Mishkin, Pamela and Clark, Jack and others},
  booktitle={International conference on machine learning},
  pages={8748--8763},
  year={2021},
  organization={PmLR}
}

@inproceedings{chen2024internvl,
  title={Internvl: Scaling up vision foundation models and aligning for generic visual-linguistic tasks},
  author={Chen, Zhe and Wu, Jiannan and Wang, Wenhai and Su, Weijie and Chen, Guo and Xing, Sen and Zhong, Muyan and Zhang, Qinglong and Zhu, Xizhou and Lu, Lewei and others},
  booktitle={Proceedings of the IEEE/CVF conference on computer vision and pattern recognition},
  pages={24185--24198},
  year={2024}
}

@inproceedings{hokamp2017lexically,
  title={Lexically Constrained Decoding for Sequence Generation Using Grid Beam Search},
  author={Hokamp, Chris and Liu, Qun},
  booktitle={Proceedings of the 55th Annual Meeting of the Association for Computational Linguistics (Volume 1: Long Papers)},
  pages={1535--1546},
  year={2017}
}

@article{tschannen2025siglip,
  title={Siglip 2: Multilingual vision-language encoders with improved semantic understanding, localization, and dense features},
  author={Tschannen, Michael and Gritsenko, Alexey and Wang, Xiao and Naeem, Muhammad Ferjad and Alabdulmohsin, Ibrahim and Parthasarathy, Nikhil and Evans, Talfan and Beyer, Lucas and Xia, Ye and Mustafa, Basil and others},
  journal={arXiv preprint arXiv:2502.14786},
  year={2025}
}

@article{dehghani2023patch,
  title={Patch n’pack: Navit, a vision transformer for any aspect ratio and resolution},
  author={Dehghani, Mostafa and Mustafa, Basil and Djolonga, Josip and Heek, Jonathan and Minderer, Matthias and Caron, Mathilde and Steiner, Andreas and Puigcerver, Joan and Geirhos, Robert and Alabdulmohsin, Ibrahim M and others},
  journal={Advances in Neural Information Processing Systems},
  volume={36},
  pages={2252--2274},
  year={2023}
}

@article{qwen2025qwen25tech,
      title={Qwen2.5 Technical Report}, 
      author={Qwen and : and An Yang and Baosong Yang and Beichen Zhang and Binyuan Hui and Bo Zheng and Bowen Yu and Chengyuan Li and Dayiheng Liu and Fei Huang and Haoran Wei and Huan Lin and Jian Yang and Jianhong Tu and Jianwei Zhang and Jianxin Yang and Jiaxi Yang and Jingren Zhou and Junyang Lin and Kai Dang and Keming Lu and Keqin Bao and Kexin Yang and Le Yu and Mei Li and Mingfeng Xue and Pei Zhang and Qin Zhu and Rui Men and Runji Lin and Tianhao Li and Tianyi Tang and Tingyu Xia and Xingzhang Ren and Xuancheng Ren and Yang Fan and Yang Su and Yichang Zhang and Yu Wan and Yuqiong Liu and Zeyu Cui and Zhenru Zhang and Zihan Qiu},
      journal={arXiv preprint arXiv:2412.15115},
      year={2025}
}

@misc{flux2024,
    author={Black Forest Labs},
    title={FLUX},
    year={2024},
    howpublished={\url{https://github.com/black-forest-labs/flux}},
}

@article{labs2025flux,
  title={FLUX. 1 Kontext: Flow Matching for In-Context Image Generation and Editing in Latent Space},
  author={Labs, Black Forest and Batifol, Stephen and Blattmann, Andreas and Boesel, Frederic and Consul, Saksham and Diagne, Cyril and Dockhorn, Tim and English, Jack and English, Zion and Esser, Patrick and others},
  journal={arXiv preprint arXiv:2506.15742},
  year={2025}
}

@inproceedings{gao2025vinabench,
  title={Vinabench: Benchmark for faithful and consistent visual narratives},
  author={Gao, Silin and Mathew, Sheryl and Mi, Li and Mamooler, Sepideh and Zhao, Mengjie and Wakaki, Hiromi and Mitsufuji, Yuki and Montariol, Syrielle and Bosselut, Antoine},
  booktitle={Proceedings of the Computer Vision and Pattern Recognition Conference},
  pages={2870--2879},
  year={2025}
}

@inproceedings{lai2024lego,
  title={LEGO: L earning EGO centric Action Frame Generation via Visual Instruction Tuning},
  author={Lai, Bolin and Dai, Xiaoliang and Chen, Lawrence and Pang, Guan and Rehg, James M and Liu, Miao},
  booktitle={European Conference on Computer Vision},
  pages={135--155},
  year={2024}
}

@inproceedings{lin2022egocentric,
  title={Egocentric video-language pretraining},
  author={Lin, Kevin Qinghong and Wang, Alex Jinpeng and Soldan, Mattia and Wray, Michael and Yan, Rui and Xu, Eric Zhongcong and Gao, Difei and Tu, Rongcheng and Zhao, Wenzhe and Kong, Weijie and others},
  booktitle={Proceedings of the 36th International Conference on Neural Information Processing Systems},
  pages={7575--7586},
  year={2022}
}

@article{heusel2017gans,
  title={Gans trained by a two time-scale update rule converge to a local nash equilibrium},
  author={Heusel, Martin and Ramsauer, Hubert and Unterthiner, Thomas and Nessler, Bernhard and Hochreiter, Sepp},
  journal={Advances in neural information processing systems},
  volume={30},
  year={2017}
}

@inproceedings{zhang2018unreasonable,
  title={The unreasonable effectiveness of deep features as a perceptual metric},
  author={Zhang, Richard and Isola, Phillip and Efros, Alexei A and Shechtman, Eli and Wang, Oliver},
  booktitle={Proceedings of the IEEE conference on computer vision and pattern recognition},
  pages={586--595},
  year={2018}
}

@article{iandola2016squeezenet,
  title={SqueezeNet: AlexNet-level accuracy with 50x fewer parameters and< 0.5 MB model size},
  author={Iandola, Forrest N and Han, Song and Moskewicz, Matthew W and Ashraf, Khalid and Dally, William J and Keutzer, Kurt},
  journal={arXiv preprint arXiv:1602.07360},
  year={2016}
}

@article{salimans2022progressive,
  title={Progressive distillation for fast sampling of diffusion models},
  author={Salimans, Tim and Ho, Jonathan},
  journal={arXiv preprint arXiv:2202.00512},
  year={2022}
}

@article{grattafiori2024llama,
  title={The llama 3 herd of models},
  author={Grattafiori, Aaron and Dubey, Abhimanyu and Jauhri, Abhinav and Pandey, Abhinav and Kadian, Abhishek and Al-Dahle, Ahmad and Letman, Aiesha and Mathur, Akhil and Schelten, Alan and Vaughan, Alex and others},
  journal={arXiv preprint arXiv:2407.21783},
  year={2024}
}

@article{achiam2023gpt,
  title={Gpt-4 technical report},
  author={Achiam, Josh and Adler, Steven and Agarwal, Sandhini and Ahmad, Lama and Akkaya, Ilge and Aleman, Florencia Leoni and Almeida, Diogo and Altenschmidt, Janko and Altman, Sam and Anadkat, Shyamal and others},
  journal={arXiv preprint arXiv:2303.08774},
  year={2023}
}

@article{hurst2024gpt,
  title={Gpt-4o system card},
  author={Hurst, Aaron and Lerer, Adam and Goucher, Adam P and Perelman, Adam and Ramesh, Aditya and Clark, Aidan and Ostrow, AJ and Welihinda, Akila and Hayes, Alan and Radford, Alec and others},
  journal={arXiv preprint arXiv:2410.21276},
  year={2024}
}

@article{singh2025openai,
  title={OpenAI GPT-5 System Card},
  author={Singh, Aaditya and Fry, Adam and Perelman, Adam and Tart, Adam and Ganesh, Adi and El-Kishky, Ahmed and McLaughlin, Aidan and Low, Aiden and Ostrow, AJ and Ananthram, Akhila and others},
  journal={arXiv preprint arXiv:2601.03267},
  year={2025}
}

@article{yang2025qwen3,
  title={Qwen3 technical report},
  author={Yang, An and Li, Anfeng and Yang, Baosong and Zhang, Beichen and Hui, Binyuan and Zheng, Bo and Yu, Bowen and Gao, Chang and Huang, Chengen and Lv, Chenxu and others},
  journal={arXiv preprint arXiv:2505.09388},
  year={2025}
}

@article{comanici2025gemini,
  title={Gemini 2.5: Pushing the frontier with advanced reasoning, multimodality, long context, and next generation agentic capabilities},
  author={Comanici, Gheorghe and Bieber, Eric and Schaekermann, Mike and Pasupat, Ice and Sachdeva, Noveen and Dhillon, Inderjit and Blistein, Marcel and Ram, Ori and Zhang, Dan and Rosen, Evan and others},
  journal={arXiv preprint arXiv:2507.06261},
  year={2025}
}

@article{qin2024worldsimbench,
  title={Worldsimbench: Towards video generation models as world simulators},
  author={Qin, Yiran and Shi, Zhelun and Yu, Jiwen and Wang, Xijun and Zhou, Enshen and Li, Lijun and Yin, Zhenfei and Liu, Xihui and Sheng, Lu and Shao, Jing and others},
  journal={arXiv preprint arXiv:2410.18072},
  year={2024}
}

@article{duan2025worldscore,
  title={Worldscore: A unified evaluation benchmark for world generation},
  author={Duan, Haoyi and Yu, Hong-Xing and Chen, Sirui and Fei-Fei, Li and Wu, Jiajun},
  journal={arXiv preprint arXiv:2504.00983},
  year={2025}
}

@article{zhuang2025vistorybench,
  title={Vistorybench: Comprehensive benchmark suite for story visualization},
  author={Zhuang, Cailin and Huang, Ailin and Cheng, Wei and Wu, Jingwei and Hu, Yaoqi and Liao, Jiaqi and Wang, Hongyuan and Liao, Xinyao and Cai, Weiwei and Xu, Hengyuan and others},
  journal={arXiv preprint arXiv:2505.24862},
  year={2025}
}

@article{wu2024conceptmix,
  title={Conceptmix: A compositional image generation benchmark with controllable difficulty},
  author={Wu, Xindi and Yu, Dingli and Huang, Yangsibo and Russakovsky, Olga and Arora, Sanjeev},
  journal={Advances in Neural Information Processing Systems},
  volume={37},
  pages={86004--86047},
  year={2024}
}

@article{alayrac2022flamingo,
  title={Flamingo: a visual language model for few-shot learning},
  author={Alayrac, Jean-Baptiste and Donahue, Jeff and Luc, Pauline and Miech, Antoine and Barr, Iain and Hasson, Yana and Lenc, Karel and Mensch, Arthur and Millican, Katherine and Reynolds, Malcolm and others},
  journal={Advances in neural information processing systems},
  volume={35},
  pages={23716--23736},
  year={2022}
}

@article{zhu2023multimodal,
  title={Multimodal c4: An open, billion-scale corpus of images interleaved with text},
  author={Zhu, Wanrong and Hessel, Jack and Awadalla, Anas and Gadre, Samir Yitzhak and Dodge, Jesse and Fang, Alex and Yu, Youngjae and Schmidt, Ludwig and Wang, William Yang and Choi, Yejin},
  journal={Advances in Neural Information Processing Systems},
  volume={36},
  pages={8958--8974},
  year={2023}
}

@article{laurenccon2023obelics,
  title={Obelics: An open web-scale filtered dataset of interleaved image-text documents},
  author={Lauren{\c{c}}on, Hugo and Saulnier, Lucile and Tronchon, L{\'e}o and Bekman, Stas and Singh, Amanpreet and Lozhkov, Anton and Wang, Thomas and Karamcheti, Siddharth and Rush, Alexander and Kiela, Douwe and others},
  journal={Advances in Neural Information Processing Systems},
  volume={36},
  pages={71683--71702},
  year={2023}
}

@inproceedings{sun2024emu,
  title={Emu: Generative Pretraining in Multimodality},
  author={Sun, Quan and Yu, Qiying and Cui, Yufeng and Zhang, Fan and Zhang, Xiaosong and Wang, Yueze and Gao, Hongcheng and Liu, Jingjing and Huang, Tiejun and Wang, Xinlong},
  booktitle={The Twelfth International Conference on Learning Representations},
  year={2024}
}

@article{tian2024mm,
  title={Mm-interleaved: Interleaved image-text generative modeling via multi-modal feature synchronizer},
  author={Tian, Changyao and Zhu, Xizhou and Xiong, Yuwen and Wang, Weiyun and Chen, Zhe and Wang, Wenhai and Chen, Yuntao and Lu, Lewei and Lu, Tong and Zhou, Jie and others},
  journal={arXiv preprint arXiv:2401.10208},
  year={2024}
}

@article{team2024chameleon,
  title={Chameleon: Mixed-modal early-fusion foundation models},
  author={Team, Chameleon},
  journal={arXiv preprint arXiv:2405.09818},
  year={2024}
}

@inproceedings{sun2024generative,
  title={Generative multimodal models are in-context learners},
  author={Sun, Quan and Cui, Yufeng and Zhang, Xiaosong and Zhang, Fan and Yu, Qiying and Wang, Yueze and Rao, Yongming and Liu, Jingjing and Huang, Tiejun and Wang, Xinlong},
  booktitle={Proceedings of the IEEE/CVF Conference on Computer Vision and Pattern Recognition},
  pages={14398--14409},
  year={2024}
}

@inproceedings{dong2024dreamllm,
  title={DreamLLM: Synergistic Multimodal Comprehension and Creation},
  author={Dong, Runpei and Han, Chunrui and Peng, Yuang and Qi, Zekun and Ge, Zheng and Yang, Jinrong and Zhao, Liang and Sun, Jianjian and Zhou, Hongyu and Wei, Haoran and others},
  booktitle={The Twelfth International Conference on Learning Representations},
  year={2024}
}

@article{chern2024anole,
  title={Anole: An open, autoregressive, native large multimodal models for interleaved image-text generation},
  author={Chern, Ethan and Su, Jiadi and Ma, Yan and Liu, Pengfei},
  journal={arXiv preprint arXiv:2407.06135},
  year={2024}
}

@inproceedings{lin2024learning,
  title={Learning Video Context as Interleaved Multimodal Sequences},
  author={Lin, Kevin Qinghong and Zhang, Pengchuan and Gao, Difei and Xia, Xide and Chen, Joya and Gao, Ziteng and Xie, Jinheng and Xiao, Xuhong and Shou, Mike Zheng},
  booktitle={European Conference on Computer Vision},
  pages={375--396},
  year={2024}
}

@article{qu2025eo,
  title={Eo-1: Interleaved vision-text-action pretraining for general robot control},
  author={Qu, Delin and Song, Haoming and Chen, Qizhi and Chen, Zhaoqing and Gao, Xianqiang and Ye, Xinyi and Lv, Qi and Shi, Modi and Ren, Guanghui and Ruan, Cheng and others},
  journal={arXiv preprint arXiv:2508.21112},
  year={2025}
}

@article{zhang2025matrix,
  title={Matrix-Game: Interactive World Foundation Model},
  author={Zhang, Yifan and Peng, Chunli and Wang, Boyang and Wang, Puyi and Zhu, Qingcheng and Kang, Fei and Jiang, Biao and Gao, Zedong and Li, Eric and Liu, Yang and others},
  journal={arXiv preprint arXiv:2506.18701},
  year={2025}
}

@article{he2025matrix,
  title={Matrix-game 2.0: An open-source real-time and streaming interactive world model},
  author={He, Xianglong and Peng, Chunli and Liu, Zexiang and Wang, Boyang and Zhang, Yifan and Cui, Qi and Kang, Fei and Jiang, Biao and An, Mengyin and Ren, Yangyang and others},
  journal={arXiv preprint arXiv:2508.13009},
  year={2025}
}

@inproceedings{lin2024evaluating,
  title={Evaluating Text-to-Visual Generation with Image-to-Text Generation},
  author={Lin, Zhiqiu and Pathak, Deepak and Li, Baiqi and Li, Jiayao and Xia, Xide and Neubig, Graham and Zhang, Pengchuan and Ramanan, Deva},
  booktitle={European Conference on Computer Vision},
  pages={366--384},
  year={2024}
}

@article{ding2025understanding,
  title={Understanding world or predicting future? a comprehensive survey of world models},
  author={Ding, Jingtao and Zhang, Yunke and Shang, Yu and Zhang, Yuheng and Zong, Zefang and Feng, Jie and Yuan, Yuan and Su, Hongyuan and Li, Nian and Sukiennik, Nicholas and others},
  journal={ACM Computing Surveys},
  volume={58},
  number={3},
  pages={1--38},
  year={2025},
  publisher={ACM New York, NY}
}

@article{ha2018world,
  title={World Models},
  author={Ha, David and Schmidhuber, J{\"u}rgen},
  journal={arXiv preprint arXiv:1803.10122},
  year={2018}
}

@inproceedings{gurnee2024language,
  title={Language Models Represent Space and Time},
  author={Wes Gurnee and Max Tegmark},
  booktitle={The Twelfth International Conference on Learning Representations},
  year={2024}
}

@article{lecun2022path,
  title={A path towards autonomous machine intelligence},
  author={LeCun, Yann},
  journal={OpenReview},
  volume={62},
  year={2022}
}

@inproceedings{hao2023reasoning,
  title={Reasoning with language model is planning with world model},
  author={Hao, Shibo and Gu, Yi and Ma, Haodi and Hong, Joshua and Wang, Zhen and Wang, Daisy and Hu, Zhiting},
  booktitle={Proceedings of the 2023 Conference on Empirical Methods in Natural Language Processing},
  pages={8154--8173},
  year={2023}
}

@inproceedings{yang2023learning,
  title={Learning Interactive Real-World Simulators},
  author={Yang, Sherry and Du, Yilun and Ghasemipour, Seyed Kamyar Seyed and Tompson, Jonathan and Schuurmans, Dale and Abbeel, Pieter},
  booktitle={NeurIPS 2023 Workshop on Generalization in Planning},
  year={2023}
}

@article{liu2024sora,
  title={Sora: A review on background, technology, limitations, and opportunities of large vision models},
  author={Liu, Yixin and Zhang, Kai and Li, Yuan and Yan, Zhiling and Gao, Chujie and Chen, Ruoxi and Yuan, Zhengqing and Huang, Yue and Sun, Hanchi and Gao, Jianfeng and others},
  journal={arXiv preprint arXiv:2402.17177},
  year={2024}
}

@inproceedings{wu2023daydreamer,
  title={Daydreamer: World models for physical robot learning},
  author={Wu, Philipp and Escontrela, Alejandro and Hafner, Danijar and Abbeel, Pieter and Goldberg, Ken},
  booktitle={Conference on robot learning},
  pages={2226--2240},
  year={2023},
  organization={PMLR}
}

@inproceedings{park2023generative,
  title={Generative agents: Interactive simulacra of human behavior},
  author={Park, Joon Sung and O'Brien, Joseph and Cai, Carrie Jun and Morris, Meredith Ringel and Liang, Percy and Bernstein, Michael S},
  booktitle={Proceedings of the 36th annual acm symposium on user interface software and technology},
  pages={1--22},
  year={2023}
}

@article{gao2024vista,
  title={Vista: A generalizable driving world model with high fidelity and versatile controllability},
  author={Gao, Shenyuan and Yang, Jiazhi and Chen, Li and Chitta, Kashyap and Qiu, Yihang and Geiger, Andreas and Zhang, Jun and Li, Hongyang},
  journal={Advances in Neural Information Processing Systems},
  volume={37},
  pages={91560--91596},
  year={2024}
}

@article{pittore2000learning,
  title={Learning to Recognize Visual Dynamic Events from Examples},
  author={Pittore, Massimiliano and Campani, Marco and Verri, Alessandro},
  journal={International Journal of Computer Vision},
  volume={38},
  number={1},
  pages={35--44},
  year={2000},
  publisher={Kluwer Academic Publishers Norwell, MA, USA}
}

@article{buxton2003learning,
  title={Learning and understanding dynamic scene activity: a review},
  author={Buxton, Hilary},
  journal={Image and vision computing},
  volume={21},
  number={1},
  pages={125--136},
  year={2003},
  publisher={Elsevier}
}

@article{nienhaus2005depicting,
  title={Depicting dynamics using principles of visual art and narrations},
  author={Nienhaus, Marc and Dollner, Jurgen},
  journal={IEEE Computer Graphics and Applications},
  volume={25},
  number={3},
  pages={40--51},
  year={2005},
  publisher={IEEE}
}

@article{xue2016visual,
  title={Visual dynamics: Probabilistic future frame synthesis via cross convolutional networks},
  author={Xue, Tianfan and Wu, Jiajun and Bouman, Katherine and Freeman, Bill},
  journal={Advances in neural information processing systems},
  volume={29},
  year={2016}
}

@article{finn2016unsupervised,
  title={Unsupervised learning for physical interaction through video prediction},
  author={Finn, Chelsea and Goodfellow, Ian and Levine, Sergey},
  journal={Advances in neural information processing systems},
  volume={29},
  year={2016}
}

@article{gepshtein2000emergence,
  title={The emergence of visual objects in space--time},
  author={Gepshtein, Sergei and Kubovy, Michael},
  journal={Proceedings of the National Academy of Sciences},
  volume={97},
  number={14},
  pages={8186--8191},
  year={2000},
  publisher={The National Academy of Sciences}
}

@book{dickmanns2007dynamic,
  title={Dynamic vision for perception and control of motion},
  author={Dickmanns, Ernst D},
  year={2007},
  publisher={Springer}
}

@article{johansson1975visual,
  title={Visual motion perception},
  author={Johansson, Gunnar},
  journal={Scientific American},
  volume={232},
  number={6},
  pages={76--89},
  year={1975},
  publisher={JSTOR}
}

@article{bonnet2005visual,
  title={Visual representations of dynamic actions from static pictures},
  author={Bonnet, Claude and Paulos, Carlos and Nithart, Christelle},
  journal={Perception},
  volume={34},
  number={7},
  pages={835--846},
  year={2005},
  publisher={SAGE Publications Sage UK: London, England}
}

@inproceedings{rowe1995statistical,
  title={Statistical background modelling for tracking with a virtual camera},
  author={Rowe, Simon and Blake, Andrew},
  booktitle={Proceedings of the 6th British conference on Machine vision (Vol. 2)},
  pages={423--432},
  year={1995}
}

@article{yao2025beyond,
  title={Beyond Static Scenes: Camera-controllable Background Generation for Human Motion},
  author={Yao, Mingshuai and Chen, Mengting and Zhou, Qinye and Zhang, Yabo and Liu, Ming and Li, Xiaoming and Liu, Shaohui and Ju, Chen and Xiao, Shuai and Liu, Qingwen and others},
  journal={arXiv preprint arXiv:2504.02004},
  year={2025}
}

@inproceedings{soldan2022mad,
  title={Mad: A scalable dataset for language grounding in videos from movie audio descriptions},
  author={Soldan, Mattia and Pardo, Alejandro and Alc{\'a}zar, Juan Le{\'o}n and Caba, Fabian and Zhao, Chen and Giancola, Silvio and Ghanem, Bernard},
  booktitle={Proceedings of the IEEE/CVF Conference on Computer Vision and Pattern Recognition},
  pages={5026--5035},
  year={2022}
}

@article{fan2022minedojo,
  title={Minedojo: Building open-ended embodied agents with internet-scale knowledge},
  author={Fan, Linxi and Wang, Guanzhi and Jiang, Yunfan and Mandlekar, Ajay and Yang, Yuncong and Zhu, Haoyi and Tang, Andrew and Huang, De-An and Zhu, Yuke and Anandkumar, Anima},
  journal={Advances in Neural Information Processing Systems},
  volume={35},
  pages={18343--18362},
  year={2022}
}

@inproceedings{wang2025koala,
  title={Koala-36m: A large-scale video dataset improving consistency between fine-grained conditions and video content},
  author={Wang, Qiuheng and Shi, Yukai and Ou, Jiarong and Chen, Rui and Lin, Ke and Wang, Jiahao and Jiang, Boyuan and Yang, Haotian and Zheng, Mingwu and Tao, Xin and others},
  booktitle={Proceedings of the Computer Vision and Pattern Recognition Conference},
  pages={8428--8437},
  year={2025}
}

@article{dinkevich2025story2board,
  title={Story2board: A training-free approach for expressive storyboard generation},
  author={Dinkevich, David and Levy, Matan and Avrahami, Omri and Samuel, Dvir and Lischinski, Dani},
  journal={arXiv preprint arXiv:2508.09983},
  year={2025}
}

@inproceedings{pan2024synthesizing,
  title={Synthesizing coherent story with auto-regressive latent diffusion models},
  author={Pan, Xichen and Qin, Pengda and Li, Yuhong and Xue, Hui and Chen, Wenhu},
  booktitle={Proceedings of the IEEE/CVF Winter Conference on Applications of Computer Vision},
  pages={2920--2930},
  year={2024}
}

@inproceedings{liu2024intelligent,
  title={Intelligent Grimm-Open-ended Visual Storytelling via Latent Diffusion Models},
  author={Liu, Chang and Wu, Haoning and Zhong, Yujie and Zhang, Xiaoyun and Wang, Yanfeng and Xie, Weidi},
  booktitle={Proceedings of the IEEE/CVF Conference on Computer Vision and Pattern Recognition},
  pages={6190--6200},
  year={2024}
}

@article{team2025gemma,
  title={Gemma 3 technical report},
  author={Team, Gemma and Kamath, Aishwarya and Ferret, Johan and Pathak, Shreya and Vieillard, Nino and Merhej, Ramona and Perrin, Sarah and Matejovicova, Tatiana and Ram{\'e}, Alexandre and Rivi{\`e}re, Morgane and others},
  journal={arXiv preprint arXiv:2503.19786},
  year={2025}
}

@article{dosovitskiy2020image,
  title={An image is worth 16x16 words: Transformers for image recognition at scale},
  author={Dosovitskiy, Alexey},
  journal={arXiv preprint arXiv:2010.11929},
  year={2020}
}

@inproceedings{islam2024video,
  title={Video recap: Recursive captioning of hour-long videos},
  author={Islam, Md Mohaiminul and Ho, Ngan and Yang, Xitong and Nagarajan, Tushar and Torresani, Lorenzo and Bertasius, Gedas},
  booktitle={Proceedings of the IEEE/CVF Conference on Computer Vision and Pattern Recognition},
  pages={18198--18208},
  year={2024}
}

@inproceedings{rombach2022high,
  title={High-resolution image synthesis with latent diffusion models},
  author={Rombach, Robin and Blattmann, Andreas and Lorenz, Dominik and Esser, Patrick and Ommer, Bj{\"o}rn},
  booktitle={Proceedings of the IEEE/CVF conference on computer vision and pattern recognition},
  pages={10684--10695},
  year={2022}
}

@inproceedings{li2022blip,
  title={Blip: Bootstrapping language-image pre-training for unified vision-language understanding and generation},
  author={Li, Junnan and Li, Dongxu and Xiong, Caiming and Hoi, Steven},
  booktitle={International conference on machine learning},
  pages={12888--12900},
  year={2022},
  organization={PMLR}
}

@article{zheng2023judging,
  title={Judging llm-as-a-judge with mt-bench and chatbot arena},
  author={Zheng, Lianmin and Chiang, Wei-Lin and Sheng, Ying and Zhuang, Siyuan and Wu, Zhanghao and Zhuang, Yonghao and Lin, Zi and Li, Zhuohan and Li, Dacheng and Xing, Eric and others},
  journal={Advances in Neural Information Processing Systems},
  volume={36},
  pages={46595--46623},
  year={2023}
}

@article{zhu2020deformable,
  title={Deformable detr: Deformable transformers for end-to-end object detection},
  author={Zhu, Xizhou and Su, Weijie and Lu, Lewei and Li, Bin and Wang, Xiaogang and Dai, Jifeng},
  journal={arXiv preprint arXiv:2010.04159},
  year={2020}
}

@inproceedings{kwon2023efficient,
  title={Efficient memory management for large language model serving with pagedattention},
  author={Kwon, Woosuk and Li, Zhuohan and Zhuang, Siyuan and Sheng, Ying and Zheng, Lianmin and Yu, Cody Hao and Gonzalez, Joseph and Zhang, Hao and Stoica, Ion},
  booktitle={Proceedings of the 29th symposium on operating systems principles},
  pages={611--626},
  year={2023}
}

@inproceedings{loshchilov2018decoupled,
  title={Decoupled Weight Decay Regularization},
  author={Loshchilov, Ilya and Hutter, Frank},
  booktitle={International Conference on Learning Representations},
  year={2018}
}

@article{li2024llava,
  title={Llava-onevision: Easy visual task transfer},
  author={Li, Bo and Zhang, Yuanhan and Guo, Dong and Zhang, Renrui and Li, Feng and Zhang, Hao and Zhang, Kaichen and Zhang, Peiyuan and Li, Yanwei and Liu, Ziwei and others},
  journal={arXiv preprint arXiv:2408.03326},
  year={2024}
}

@article{han2023improving,
  title={Improving tuning-free real image editing with proximal guidance},
  author={Han, Ligong and Wen, Song and Chen, Qi and Zhang, Zhixing and Song, Kunpeng and Ren, Mengwei and Gao, Ruijiang and Stathopoulos, Anastasis and He, Xiaoxiao and Chen, Yuxiao and others},
  journal={arXiv preprint arXiv:2306.05414},
  year={2023}
}

@article{meng2021sdedit,
  title={Sdedit: Guided image synthesis and editing with stochastic differential equations},
  author={Meng, Chenlin and He, Yutong and Song, Yang and Song, Jiaming and Wu, Jiajun and Zhu, Jun-Yan and Ermon, Stefano},
  journal={arXiv preprint arXiv:2108.01073},
  year={2021}
}

@inproceedings{brooks2023instructpix2pix,
  title={Instructpix2pix: Learning to follow image editing instructions},
  author={Brooks, Tim and Holynski, Aleksander and Efros, Alexei A},
  booktitle={Proceedings of the IEEE/CVF conference on computer vision and pattern recognition},
  pages={18392--18402},
  year={2023}
}
\bibliographystyle{icml2026}

\newpage
\appendix
\onecolumn

\section{\vwm{} Implementation Details}
\label{sec:pt_appdx}

\subsection{State-Transition Data Construction Details}
\label{sec:data_cons_appdx}

\paragraph{State Extraction}
We extract our world modeling states as keyframes from three source video datasets, including Ego4D \citep{grauman2022ego4d}, AgiBotWorld-Alpha \citep{bu2025agibot} and five sub-portions (Panda-70M, Pexels, Pixabay, Mixkit, BDD100K) of ShareGPT4Video \citep{chen2024sharegpt4video}.
We control the number of videos selected from each source dataset based on a balanced number of extracted states ($\sim53K$ states from each source).
Note that videos from Ego4D are mostly very long, which can reach to $30$ minutes, resulting in a large number of states extracted from just a single Ego4D video.
To include more diverse Ego4D videos, while maintain their number of extracted states to be comparable with the other two sources, we sample a sub-segment of each Ego4D video (instead of the whole video) to perform the state extraction, according to the video segmentation annotated by Ego4D-HCap \citep{islam2024video}.
Table~\ref{tab:state_data_stats_appdx} includes the detailed descriptions of each source dataset (or sub-portion) and the statistics of our state extraction.

\paragraph{Transition Annotation}
We prompt an oracle vision language model (VLM), InternVL-78B-Instruct \citep{chen2024internvl}, to annotate the transition between each pair of adjacent states (or keyframes) extracted from our source videos.
InternVL is first instructed to generate a dense caption of the former state, to understand the visual contents presented at the start of the transition.
Based on that, InternVL is then prompted by a fine-grained task demonstration to comprehensively analyze the transition following our defined schema in Sec~\ref{sec_data}.
The prompts for dense caption and transition analysis are below:
\begin{tcolorbox}[
    colback=white,
    colframe=black!60,
    title=\textbf{Oracle VLM Prompt for Dense Caption},
    coltitle=white,
    colbacktitle=black!60,
    fonttitle=\bfseries,
    sharp corners,
    boxrule=0.8pt,
    left=8pt,
    right=8pt,
    top=8pt,
    bottom=8pt
]
You are given an image. Your goal is to provide a concise but detailed caption for the image.
Your caption should address all the objects in the scene and their spatial relationships to each other.\par
\vspace{1.0em}
Follow these quality standards:\par
\vspace{0.5em}
Precision: Use specific, measurable descriptions rather than vague qualifiers.\par
Accuracy: Ensure all observations are factually correct and verifiable through direct visual inspection. Do not use suggestive language about the image (e.g.``suggesting a point-of-view angle...'' or ``indicating that...''). Do not describe anything beyond what is immediately provided in the image.\par
Evidence-Based Grounding: Base every claim on observable pixel-level evidence in the frames - cite specific visual details, colors, shapes, positions, and changes that support your analysis.\par
Completeness: Address all relevant aspects of the scene.\par
Consistency: Maintain consistent terminology and granularity throughout the analysis.\par
Relevance: Focus on description that meaningfully contributes to the task while being concise. If you are unsure of what a specific object in the scene is, avoid referring to it.\par
Clarity: Ensure descriptions are unambiguous and interpretable by other systems.\par
\vspace{1.0em}
Final Output Format:\par
Enclose your caption within a structured JSON output following this exact schema:\par
\{\par
\quad``type'': ``object'',\par
\quad``properties'': \{\par
\quad\quad``caption'': \{``type'': ``string''\}\par
\quad\}\par
\}\par
\vspace{1.0em}
Remember: Your goal is to provide a concise, yet detailed caption for the image you are given. Your caption must be specific and factually correct, containing no erroneous statements about the image.\par
Do not describe anything beyond what is immediately provided in the image. Everything must be factually correct and verifiable from the image. Be concise. If you are unsure about something, simply do not say it.\par
\vspace{1.0em}
Here is the image: $\langle$the former state image$\rangle$
\end{tcolorbox}

\begin{table*}[t]
\centering
\caption{Detailed statistics of state extraction in \vwm{} training data, including number of videos (\textbf{\# Videos}) selected from each source dataset (or sub-portion), average (\textbf{Avg.}), minimum (\textbf{Min}) and maximum (\textbf{Max}) \textbf{Duration} of a video in seconds, total number of states (\textbf{\# States}) extracted from each source dataset, average, minimum and maximum number of states extracted from per video (\textbf{States per Video}), average, minimum and maximum time interval in seconds between each pair of adjacent states (\textbf{State Interval}).}
\resizebox{1.0\textwidth}{!}{
\smallskip\begin{tabular}{@{~}lll@{~~~}cc@{~~~}c@{~~~}ccc@{~~~}c@{~~~}cc@{~~~}c@{~~~}c@{~}}
\toprule
\multirow{2}*{\textbf{Source}} & \multirow{2}*{\textbf{Portion}} & \multirow{2}*{\textbf{Domain}} & \multirow{2}*{\textbf{\# Videos}} & \multicolumn{3}{c}{\textbf{Duration (s)}} & \multirow{2}*{\textbf{\# States}} & \multicolumn{3}{c}{\textbf{States per Video}} & \multicolumn{3}{c}{\textbf{State Interval (s)}} \\
\cmidrule(lr){5-7}  \cmidrule(lr){9-11}  \cmidrule(lr){12-14}
& & &  & Avg. & Min & Max &  & Avg. & Min & Max & Avg. & Min & Max \\
\toprule
 \multirow{2}*{Ego4D(-HCap)} & \multirow{2}*{-} & ego-centric human daily life (household, & \multirow{2}*{1000} & \multirow{2}*{117.06} & \multirow{2}*{2.10} & \multirow{2}*{309.90} & \multirow{2}*{52976} & \multirow{2}*{52.98} & \multirow{2}*{2} & \multirow{2}*{100} & \multirow{2}*{2.25} & \multirow{2}*{0.80} & \multirow{2}*{167.10} \\
  &  & outdoor, workplace, leisure, etc.) activities &  &  &  &  &  &  \\
\midrule
 \multirow{2}*{AgiBotWorld-Alpha} & \multirow{2}*{-} & robot-view trajectories in domestic, retail, & \multirow{2}*{3400} & \multirow{2}*{46.09} & \multirow{2}*{1.07} & \multirow{2}*{80.07} & \multirow{2}*{53282} & \multirow{2}*{15.67} & \multirow{2}*{2} & \multirow{2}*{60} & \multirow{2}*{3.14} & \multirow{2}*{0.80} & \multirow{2}*{74.10} \\
  &  & industrial, restaurant, and office environment &  &  &  &  &  &  \\
\midrule
 \multirow{7}*{ShareGPT4Video} & Panda-70M & public YouTube videos (gaming, TV shows, \textit{etc.}) & 1200 & 23.97 & 0.90 & 49.15 & 13215 & 11.01 & 2 & 39 & 2.39 & 0.80 & 45.05 \\
    \cmidrule(lr){2-3}
  & Pexels & \multirow{2}*{aesthetically appealing stock videos} & 1800 & 19.12 & 0.83 & 274.08 & 12871 & 7.15 & 2 & 38 & 3.11 & 0.80 & 56.06 \\
  & Pixabay & \multirow{2}*{(natural landscape, cultural scenery,\textit{etc.})} & 1800 & 20.28 & 0.83 & 109.0 & 11999 & 6.67 & 2 & 40 & 3.58 & 0.80 & 58.10 \\
  & Mixkit & & 572 & 13.84 & 0.96 & 67.07 & 2248 & 3.93 & 2 & 24 & 4.72 & 0.84 & 32.03 \\
   \cmidrule(lr){2-3}
  & BDD100K & auto-driving video recordings & 603 & 36.32 & 3.02 & 41.93 & 13015 & 21.58 & 2 & 40 & 1.76 & 0.80 & 39.87 \\
   \cmidrule(lr){2-3}
  & All & - & 5975 & 21.67 & 0.83 & 274.08 & 53348 & 8.93 & 2 & 40 & 2.73 & 0.80 & 58.10 \\
\bottomrule
\end{tabular}
}
\label{tab:state_data_stats_appdx}
\end{table*}

\begin{tcolorbox}[
    colback=white,
    colframe=black!60,
    title=\textbf{Oracle VLM Prompt for Transition Analysis (Part 1)},
    coltitle=white,
    colbacktitle=black!60,
    fonttitle=\bfseries,
    sharp corners,
    boxrule=0.8pt,
    left=8pt,
    right=8pt,
    top=8pt,
    bottom=8pt
]
Your core objective is to generate comprehensive descriptions of a visual transition that ACCURATELY captures the dynamic transformation between two consecutive video frames, focusing on fine-grained visual features and their temporal relationships.
You should precisely capture what happens in the transition between two frames, only reporting verifiable descriptions you are entirely sure about and the frames give you clear evidence for.
Do not extrapolate hypothetical possibilities based on what you infer from the frames, only report exactly what you see. Ensure that all observations you make are factually correct and verifiable through direct visual inspection.\par
\vspace{0.5em}
Analysis Framework\par
\vspace{0.5em}
Step 1: High-Level Activity Identification\par
\vspace{0.2em}
First, identify the overarching activity or scenario occurring between the two frames. This should be a broad categorization that encompasses the primary action or event taking place.
Examples: ``Person preparing a meal in kitchen'', ``Vehicle navigating through traffic'', ``Athletes competing in a sports event'', ``Construction work in progress''.\par
\vspace{0.5em}
Step 2: Sub-Activity Decomposition\par
\vspace{0.2em}
Break down the high-level activity into constituent sub-activities.
These are intermediate-level actions that contribute to the overall activity.
Only list sub-activities that are visibly occurring in the given frames.
Do not list any sub-activities that you do not have clear evidence for.
You must NOT invent any sub-activities that are not immediately verifiable from the frames -- for example, if there is a knife and a carrot in the scene, unless the image clearly shows that the knife is being used to chop the carrot, you cannot say anything about the knife potentially being used to chop the carrot.
Examples for ``Person preparing a meal'': ``Gathering ingredients from pantry'', ``Chopping vegetables on cutting board'', ``Heating pan on stove'', ``Combining ingredients in bowl''.\par
\vspace{0.5em}
Step 3: Atomic Action Identification\par
\vspace{0.2em}
For each sub-activity, identify the atomic actions - the smallest meaningful units of action that cannot be further decomposed while maintaining semantic meaning. 
Only list atomic actions that actually occur in between the frames!
For example, if a bell pepper is not actually lifted or moved, DO NOT say anything about the bell pepper.
Examples for ``Chopping vegetables on cutting board'': ``Positioning knife above carrot'', ``Applying downward pressure to slice'', ``Lifting knife blade'', ``Sliding carrot segment aside''.\par
\vspace{0.5em}
Step 4: Transformation Analysis\par
\vspace{0.2em}
For each atomic action, systematically analyze the following six categories of transformations:\par
\vspace{0.2em}
4.1 Object Transformations\par
\vspace{0.2em}
New Objects Introduced: Identify objects that appear in Frame 2 but were not present or visible in Frame 1.\par
Objects Removed: Identify objects that disappear, become occluded, or move out of frame.\par
Object Persistence: Note objects that remain present but may undergo other changes.\par
\end{tcolorbox}

\begin{tcolorbox}[
    colback=white,
    colframe=black!60,
    title=\textbf{Oracle VLM Prompt for Transition Analysis (Part 2)},
    coltitle=white,
    colbacktitle=black!60,
    fonttitle=\bfseries,
    sharp corners,
    boxrule=0.8pt,
    left=8pt,
    right=8pt,
    top=8pt,
    bottom=8pt
]
4.2 Object State Transformations\par
\vspace{0.2em}
Color Changes: Variations in hue, saturation, brightness, or lighting conditions affecting object appearance.\par
Shape Deformation: Changes in object geometry, size, orientation, or physical configuration.\par
Physical Status: Alterations in object condition (broken/intact, open/closed, full/empty, wet/dry, \textit{etc.}).\par
Texture Changes: Modifications in surface appearance or material properties.\par
\vspace{0.5em}
4.3 Spatial Relation Transformations\par
\vspace{0.2em}
Positional Relationships: Changes in relative positioning between objects (above/below, left/right, front/back).\par
Contact Relationships: Modifications in physical contact or proximity (touching/separated, inside/outside, attached/detached).\par
Alignment Changes: Shifts in object alignment, orientation, or arrangement patterns.\par
\vspace{0.5em}
4.4 Action Transformations\par
\vspace{0.2em}
Action Continuation: Ongoing actions that persist from Frame 1 to Frame 2.\par
Action Completion: Actions that conclude or reach a terminal state.\par
Action Initiation: New actions that begin in Frame 2.\par
Action Interruption: Actions that pause, stop, or are disrupted.\par
\vspace{0.5em}
4.5 Motion Transformations\par
\vspace{0.2em}
Translational Motion: Linear movement in any direction (forward, backward, sideways, up, down).\par
Rotational Motion: Spinning, turning, or rotating movement around an axis.\par
Oscillatory Motion: Back-and-forth or periodic movement patterns.\par
Deformation Motion: Changes in object shape through stretching, compression, or bending.\par
\vspace{0.5em}
4.6 Camera Transformations\par
\vspace{0.2em}
Viewpoint Changes: Modifications in camera position or angle.\par
Zoom Operations: Magnification changes (zoom in/out).\par
Pan/Tilt Movements: Horizontal or vertical camera sweeping.\par
Focus Adjustments: Changes in depth of field or focal point.\par
\vspace{0.5em}
4.7 Background Transformations\par
\vspace{0.2em}
Illumination Changes: Variations in lighting conditions, shadows, or brightness.\par
Scene Shifts: Changes in background environment or setting.\par
Atmospheric Conditions: Modifications in weather, visibility, or environmental factors.\par
Contextual Elements: Changes in background objects or environmental details.\par
\vspace{0.5em}
Step 5: Contribution Analysis\par
\vspace{0.2em}
For the identified transformations of each atomic action, analyze and explain:\par
Atomic Action Contribution: How the transformations directly enable, facilitate, or result from the specific atomic action.\par
Sub-Activity Contribution: How the transformations support or advance the broader sub-activity goal.\par
Temporal Significance: The timing and sequence importance of the transformations within the overall activity flow.\par
\vspace{1.0em}
Quality Standards:\par
\vspace{0.2em}
Accuracy: Ensure all observations are factually correct and verifiable through direct visual inspection.\par
Evidence-Based Grounding: Base every claim on observable pixel-level evidence in the frames - cite specific visual details, colors, shapes, positions, and changes that support your analysis.\par
Precision: Use specific, measurable descriptions rather than vague qualifiers.\par
Completeness: Address all relevant transformation categories.\par
Consistency: Maintain consistent terminology and granularity throughout the analysis.\par
Relevance: Focus on transformations that meaningfully contribute to the activity progression.\par
Clarity: Ensure descriptions are unambiguous and interpretable by other systems.\par
\end{tcolorbox}

\begin{tcolorbox}[
    colback=white,
    colframe=black!60,
    title=\textbf{Oracle VLM Prompt for Transition Analysis (Part 3)},
    coltitle=white,
    colbacktitle=black!60,
    fonttitle=\bfseries,
    sharp corners,
    boxrule=0.8pt,
    left=8pt,
    right=8pt,
    top=8pt,
    bottom=8pt
]
Output Requirements\par
\vspace{0.2em}
Reasoning Process -- Think through your analysis step-by-step, only considering activities and actions you have clear evidence for, and documenting your reasoning for each level of the framework. Show your work in identifying activities, sub-activities, atomic actions, and transitions in a JSON as described below. Do not output anything other than the final JSON as described.\par
\vspace{0.2em}
Final Output Format -- Conclude your analysis with a structured JSON output following this exact schema:\par
\vspace{0.2em}
\{``type'': ``object'', ``properties'': \{\par
``high\_level\_activity'': \{``type'': ``string''\},\par
\quad``sub\_activities'': \{``type'': ``array'', ``items'': \{``type'': ``object'', ``properties'': \{``name'': \{``type'': ``string''\},\par
\quad\quad``atomic\_actions'': \{``type'': ``array'', ``items'': \{``type'': ``object'', ``properties'': \{``name'': \{``type'': ``string''\},\par
\quad\quad\quad``transformations'': \{``type'': ``object'', ``properties'': \{\par
\quad\quad\quad\quad``objects'': \{``type'': ``object'', ``properties'': \{\par
\quad\quad\quad\quad\quad``introduced'': \{``type'': ``array'', ``items'': \{``type'': ``string''\}\},\par
\quad\quad\quad\quad\quad``removed'': \{``type'': ``array'', ``items'': \{``type'': ``string''\}\},\par
\quad\quad\quad\quad\quad``persistent'': \{``type'': ``array'', ``items'': \{``type'': ``string''\}\}\}\},\par
\quad\quad\quad\quad``object\_states'': \{``type'': ``object'', ``properties'': \{\par
\quad\quad\quad\quad\quad``color\_changes'': \{``type'': ``array'', ``items'': \{``type'': ``string''\}\},\par
\quad\quad\quad\quad\quad``shape\_changes'': \{``type'': ``array'', ``items'': \{``type'': ``string''\}\},\par
\quad\quad\quad\quad\quad``physical\_status\_changes'': \{``type'': ``array'', ``items'': \{``type'': ``string''\}\},\par
\quad\quad\quad\quad\quad``texture\_changes'': \{``type'': ``array'', ``items'': \{``type'': ``string''\}\}\}\},\par
\quad\quad\quad\quad``spatial\_relations'': \{``type'': ``object'', ``properties'': \{\par
\quad\quad\quad\quad\quad``positional\_changes'': \{``type'': ``array'', ``items'': \{``type'': ``string''\}\},\par
\quad\quad\quad\quad\quad``contact\_changes'': \{``type'': ``array'', ``items'': \{``type'': ``string''\}\},\par
\quad\quad\quad\quad\quad``alignment\_changes'': \{``type'': ``array'',``items'': \{``type'': ``string''\}\}\}\},\par
\quad\quad\quad\quad``actions'': \{``type'': ``object'',``properties'': \{\par
\quad\quad\quad\quad\quad``continuing'': \{``type'': ``array'', ``items'': \{``type'': ``string''\}\},\par
\quad\quad\quad\quad\quad``completing'': \{``type'': ``array'', ``items'': \{``type'': ``string''\}\},\par
\quad\quad\quad\quad\quad``initiating'': \{``type'': ``array'', ``items'': \{``type'': ``string''\}\},\par
\quad\quad\quad\quad\quad``interrupting'': \{``type'': ``array'', ``items'': \{``type'': ``string''\}\}\}\},\par
\quad\quad\quad\quad``motion'': \{``type'': ``object'', ``properties'': \{\par
\quad\quad\quad\quad\quad``translational'': \{``type'': ``array'', ``items'': \{``type'': ``string''\}\},\par
\quad\quad\quad\quad\quad``rotational'': \{``type'': ``array'', ``items'': \{``type'': ``string''\}\},\par
\quad\quad\quad\quad\quad``oscillatory'': \{``type'': ``array'', ``items'': \{``type'': ``string''\}\},\par
\quad\quad\quad\quad\quad``deformation'': \{``type'': ``array'', ``items'': \{``type'': ``string''\}\}\}\},\par
\quad\quad\quad\quad``camera'': \{``type'': ``object'', ``properties'': \{\par
\quad\quad\quad\quad\quad``viewpoint\_changes'': \{``type'': ``array'', ``items'': \{``type'': ``string''\}\},\par
\quad\quad\quad\quad\quad``zoom\_changes'': \{``type'': ``array'', ``items'': \{``type'': ``string''\}\},\par
\quad\quad\quad\quad\quad``pan\_tilt'': \{``type'': ``array'', ``items'': \{``type'': ``string''\}\},\par
\quad\quad\quad\quad\quad``focus\_changes'': \{``type'': ``array'', ``items'': \{``type'': ``string''\}\}\}\},\par
\quad\quad\quad\quad``background'': \{``type'': ``object'', ``properties'': \{\par
\quad\quad\quad\quad\quad``illumination\_changes'': \{``type'': ``array'', ``items'': \{``type'': ``string''\}\},\par
\quad\quad\quad\quad\quad``scene\_shifts'': \{``type'': ``array'', ``items'': \{``type'': ``string''\}\},\par
\quad\quad\quad\quad\quad``atmospheric\_changes'': \{``type'': ``array'', ``items'': \{``type'': ``string''\}\},\par
\quad\quad\quad\quad\quad``contextual\_changes'': \{``type'': ``array'', ``items'': \{``type'': ``string''\}\}\}\}\}\},\par
\quad\quad\quad``contributions'': \{``type'': ``object'', ``properties'': \{\par
\quad\quad\quad\quad``to\_atomic\_action'': \{``type'': ``string''\},\par
\quad\quad\quad\quad``to\_sub\_activity'': \{``type'': ``string''\},\par
\quad\quad\quad\quad``temporal\_significance'': \{``type'': ``string''\}\}\}\par
\quad\quad\}\}\}\par
\quad\}\}\}\par
\}\}\par
\end{tcolorbox}

\begin{tcolorbox}[
    colback=white,
    colframe=black!60,
    title=\textbf{Oracle VLM Prompt for Transition Analysis (Part 4)},
    coltitle=white,
    colbacktitle=black!60,
    fonttitle=\bfseries,
    sharp corners,
    boxrule=0.8pt,
    left=8pt,
    right=8pt,
    top=8pt,
    bottom=8pt
]
Error Handling\par
\vspace{0.2em}
If certain transition categories are not applicable, do not include them in the JSON at all.\par
If the relationship between frames is unclear, state your assumptions explicitly. Try to avoid making any assumptions that are not factually correct and verifiable -- it is better to have less information that is more reliable than vice versa.\par
If objects are partially occluded, specify the degree of visibility and confidence level.\par
\vspace{0.5em}
Remember: Your goal is to create a comprehensive, structured understanding of visual change that captures both the obvious and subtle transformations occurring between the two frames.
All your observations must be factually correct and verifiable through direct visual inspection -- you must strive for reliable, accurate information ONLY.
You must base every claim on observable pixel-level evidence in the frames.\par
\vspace{1.0em}
Here is the first frame: $\langle$the former state image$\rangle$\par
To guide you, here is a caption of the first frame: $\langle$dense caption of the former state image$\rangle$\par
Here is the second frame: $\langle$the latter state image$\rangle$
\end{tcolorbox}

\begin{figure*}[t]
\centering
\includegraphics[width=1.0\textwidth]{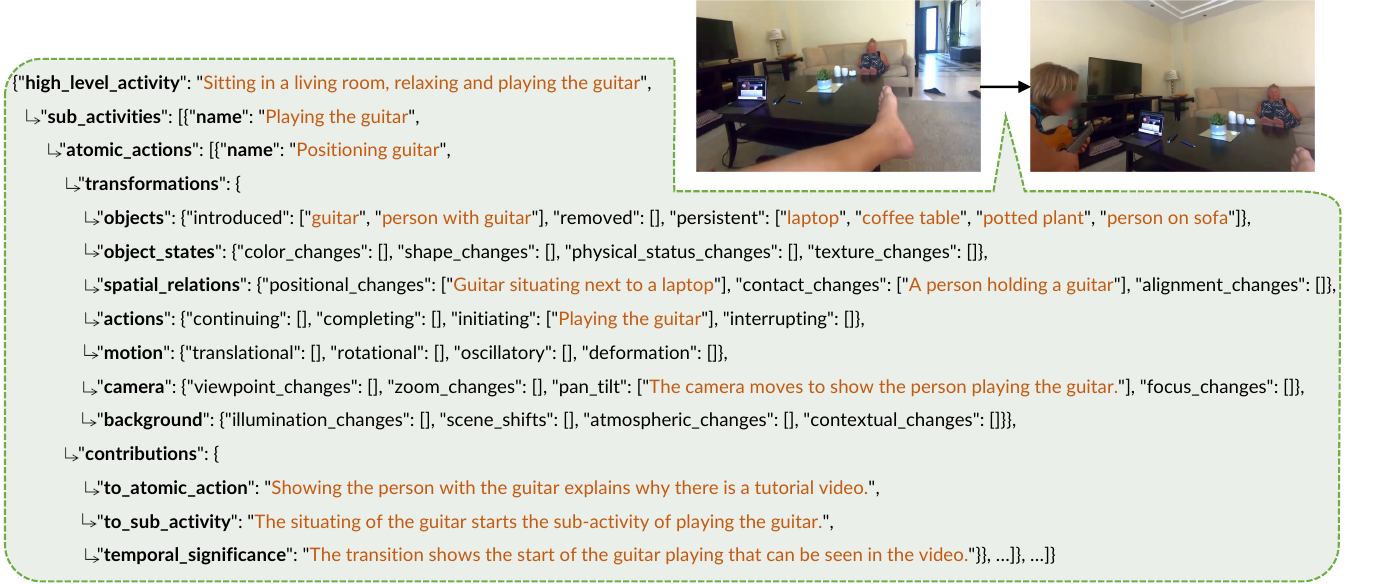}
\caption{An example of \vwm{} state transition. We frame our transition based on a hierarchical JSON schema.}
\label{data_example}
\end{figure*}

We adopt constrained decoding \citep{hokamp2017lexically} within vLLM \citep{kwon2023efficient} infrastructure to ensure that the transition generated by the oracle VLM is in valid JSON structure and follows our defined schema in above transition analysis prompt (Part 3). Figure~\ref{data_example} presents an example of our state transition annotation based on our defined JSON schema.

\subsection{\vwm{} Modeling Details}
\label{sec:model_appdx}

We initialize \vwm{} with the pre-trained weights of BAGEL-7B-MoT \citep{deng2025emerging} and conduct continued pre-training on our state-transition data.

\paragraph{Data Preprocessing}
To fit our maximum computing (GPU) memory, we split each long state-transition sequence into sub-sequences with a maximum length of $6$ states interleaved with $5$ transitions in between, where adjacent sub-sequences have an overlap of $3$ states, \textit{i.e.}, the sub-sequence sliding window has a width of $6$ states and a stride of $3$ states.
After the split, we get a total number of $48260$ state-transition (sub-)sequences as our final training samples.
To mitigate catastrophic forgetting, we further mix our state-transition data with $3000$ BAGEL’s original pre-training data samples\footnote{\url{https://github.com/ByteDance-Seed/Bagel/blob/main/TRAIN.md}}, including text-to-image generation, multi-round image-editing and LLaVA‑OneVision \citep{li2024llava} modeling samples ($1000$ each).
For each state image encoded by the VAE encoder (FLUX.1-dev\footnote{\url{https://huggingface.co/black-forest-labs/FLUX.1-dev}}, with patch size $16$), which is used for the diffusion training of state simulation, we resize the image to ensure that its longer side (height or width) is in the range of $512$ to $1024$ pixels, while keeping its original aspect ratio.
For each state image encoded by the ViT encoder (SigLIP2-so400m-patch14-384\footnote{\url{https://huggingface.co/google/siglip2-so400m-patch14-384}}, with patch size $14$), which serves as contexts for predicting subsequent transitions and states, we resize the image to ensure that its longer side fall in the range of $224$ to $518$ pixels, also under the original aspect ratio.

\paragraph{Training Hyperparameters and Computing Resources}
We train \vwm{} on a cosine learning rate scheduler, with warm-up steps, total training steps, maximum learning rate and minimum learning rate set to $1000$, $10000$, $5e^{-5}$ and $1e^{-6}$, respectively.
We deploy the training on $32$ NVIDIA GH200 GPUs, which takes about $4$ days.
To improve training efficiency, we follow BAGEL to use a dynamic batch size of training samples yielded to each GPU at each training step.
Specifically, we keep adding data samples into a batch until its total number of tokens reaches $16384$ or above, before yielding the batch to a GPU.
However, to avoid out of GPU memory, if the total number of tokens jumps directly from less than $16384$ to more than $36864$ (the maximum number of tokens allowed in a batch to fit our GPU memory) after adding a sample, we drop this sample to the next batch and iterate to add the next sample into the current batch.
Same as the BAGEL setting, we freeze the weights of the VAE encoder during the training (and the VAE decoder which is only used at the inference phase), while update the rest parts of the model.
The AdamW \citep{loshchilov2018decoupled} optimizer is used, with $\beta_{1}$, $\beta_{2}$ and $\epsilon$ set to $0.9$, $0.95$ and $1e^{-15}$, respectively, and the maximum norm for gradient clipping set to $1.0$.

\subsection{\vwm{} Validation Details}
\label{sec:valid_appdx}
We use GPT-4o \citep{hurst2024gpt,achiam2023gpt} as a VLM judge to validate \vwm{}'s performances of transition prediction and state simulation, and meanwhile to verify the quality of gold transitions and states constructed by our pipeline, based on the $900$ validation samples ($300$ each from Ego4D, AgiBotWorld-Alpha and ShareGPT4Video datasets).
The VLM judge prompts used for the transition and state validation are shown below:

\begin{tcolorbox}[
    colback=white,
    colframe=black!60,
    title=\textbf{VLM Judge Prompt for Transition Validation (Part 1)},
    coltitle=white,
    colbacktitle=black!60,
    fonttitle=\bfseries,
    sharp corners,
    boxrule=0.8pt,
    left=8pt,
    right=8pt,
    top=8pt,
    bottom=8pt
]
You are given a sequence of keyframes of a video.
Your core objective is to evaluate the reasonability of a response that predicts the dynamic transformation from the last given keyframe to the next unknown keyframe.
The response are in JSON format.
When conducting your evaluation of the response, you must focus on the following aspects:\par
\vspace{0.5em}
1. High-Level Activity: 
Whether the ``high\_level\_activity'' field reasonably predicts what will happen or continue to occur from the last given keyframe to the next keyframe?
Answer ``yes'' or ``no''.
Also output a concise justification of your answer in one or two sentences.\par
\vspace{0.5em}
2. Sub-Activities:
Whether all the ``sub\_activities'' listed in the response reasonably predict what will happen or continue to occur from the last given keyframe to the next keyframe?
Answer ``yes'' or ``no'' according to the ``name'' of each sub-activity.
Also output a concise justification of your answer in one or two sentences.\par
\vspace{0.5em}
3. Atomic Actions:
Whether all the ``atomic\_actions'' of sub-activities reasonably predict what will happen or continue to occur from the last given keyframe to the next keyframe?
Answer ``yes'' or ``no'' according to the ``name'' of each atomic action.
Also output a concise justification of your answer in one or two sentences.\par
\vspace{0.5em}
4.1. Transformations -- Objects:
Whether the ``objects'' field of the ``transformations'' reasonably predicts the objects that will be newly ``introduced'', ``removed'' or kept ``persistent'' from the last given keyframe to the next keyframe?
Answer ``empty'' if all lists in this field are empty.
Otherwise, answer ``yes'' or ``no'' according to the predictions in non-empty lists.
Also output a concise justification of your answer in one or two sentences.\par
\vspace{0.5em}
4.2. Transformations -- Object States:
Whether the ``object\_states'' field of the ``transformations'' reasonably predicts the changes of object ``color'', ``shape'', ``physical status'' or ``texture'' from the last given keyframe to the next keyframe?
Answer ``empty'' if all lists in this field are empty.
Otherwise, answer ``yes'' or ``no'' according to the predictions in non-empty lists.
Also output a concise justification of your answer in one or two sentences.\par
\end{tcolorbox}

\begin{tcolorbox}[
    colback=white,
    colframe=black!60,
    title=\textbf{VLM Judge Prompt for Transition Validation (Part 2)},
    coltitle=white,
    colbacktitle=black!60,
    fonttitle=\bfseries,
    sharp corners,
    boxrule=0.8pt,
    left=8pt,
    right=8pt,
    top=8pt,
    bottom=8pt
]
4.3. Transformations -- Spatial Relations:
Whether the ``spatial\_relations'' field of the ``transformations'' reasonably predicts the ``positional'', ``contact'' or ``alignment'' changes of objects from the last given keyframe to the next keyframe?
Answer ``empty'' if all lists in this field are empty.
Otherwise, answer ``yes'' or ``no'' according to the predictions in non-empty lists.
Also output a concise justification of your answer in one or two sentences.\par
\vspace{0.5em}
4.4. Transformations -- Actions:
Whether the ``actions'' field of the ``transformations'' reasonably predicts the ``continuing'', ``completing'', ``initiating'' or ``interrupting'' actions of objects from the last given keyframe to the next keyframe?
Answer ``empty'' if all lists in this field are empty.
Otherwise, answer ``yes'' or ``no'' according to the predictions in non-empty lists.
Also output a concise justification of your answer in one or two sentences.\par
\vspace{0.5em}
4.5. Transformations -- Motion:
Whether the ``motion'' field of the ``transformations'' reasonably predicts the ``translational'', ``rotational'', ``oscillatory'' or ``deformation'' motion of objects from the last given keyframe to the next keyframe?
Answer ``empty'' if all lists in this field are empty.
Otherwise, answer ``yes'' or ``no'' according to the predictions in non-empty lists.
Also output a concise justification of your answer in one or two sentences.\par
\vspace{0.5em}
4.6. Transformations - Camera:
Whether the ``camera'' field of the ``transformations'' reasonably predicts the ``viewpoint'', ``zoom'', ``focus'' changes or ``pan tilt'' of the camera from the last given keyframe to the next keyframe?
Answer ``empty'' if all lists in this field are empty.
Otherwise, answer ``yes'' or ``no'' according to the predictions in non-empty lists.
Also output a concise justification of your answer in one or two sentences.\par
\vspace{0.5em}
4.7. Transformations - Background:
Whether the ``background'' field of the ``transformations'' reasonably predicts the ``illumination'', ``atmospheric'', ``contextual'' changes or ``scene\_shifts'' of the background from the last given keyframe to the next keyframe?
Answer ``empty'' if all lists in this field are empty.
Otherwise, answer ``yes'' or ``no'' according to the predictions in non-empty lists.
Also output a concise justification of your answer in one or two sentences.\par
\vspace{0.5em}
5. Contributions:
Whether the ``contributions'' field reasonably analyzes the contributions of predicted ``transformations'' to ``atomic\_actions'' and ``sub\_activities'', and the ``temporal\_significance'' of predicted ``transformations''?
Answer ``yes'' or ``no''.
Also output a concise justification of your answer in one or two sentences.\par
\vspace{0.5em}
Final Output Format\par
\vspace{0.2em}
Conclude your evaluation with a structured JSON output following this exact schema:\par
\vspace{0.2em}
\{\par
``high\_level\_activity'':\{``reasonable'': ``yes'' or ``no'', ``justification'': ``string''\},\par
``sub\_activities'':\{``reasonable'': ``yes'' or ``no'', ``justification'': ``string''\},\par
``atomic\_actions'':\{``reasonable'': ``yes'' or ``no'', ``justification'': ``string''\},\par
``transformation\_objects'':\{``reasonable'': ``yes'', ``no'' or ``empty'', ``justification'': ``string''\},\par
``transformation\_object\_states'':\{``reasonable'': ``yes'', ``no'' or ``empty'', ``justification'': ``string''\},\par
``transformation\_spatial\_relations'':\{``reasonable'': ``yes'', ``no'' or ``empty'', ``justification'': ``string''\},\par
``transformation\_actions'':\{``reasonable'': ``yes'', ``no'' or ``empty'', ``justification'': ``string''\},\par
``transformation\_motion'':\{``reasonable'': ``yes'', ``no'' or ``empty'', ``justification'': ``string''\},\par
``transformation\_camera'':\{``reasonable'': ``yes'', ``no'' or ``empty'', ``justification'': ``string''\},\par
``transformation\_background'':\{``reasonable'': ``yes'', ``no'' or ``empty'', ``justification'': ``string''\},\par
``contributions'':\{``reasonable'': ``yes'' or ``no'', ``justification'': ``string''\}\par
\}\par
\vspace{0.2em}
Do not output anything other than the final JSON as described.\par
\vspace{0.5em}
Here is the sequence of video keyframes: $\langle$preceding state images before the transition to be evaluated$\rangle$\par
\vspace{0.2em}
Here is the response to evaluate: $\langle$the transition to be evaluated$\rangle$\par
\end{tcolorbox}

\begin{tcolorbox}[
    colback=white,
    colframe=black!60,
    title=\textbf{VLM Judge Prompt for State Validation},
    coltitle=white,
    colbacktitle=black!60,
    fonttitle=\bfseries,
    sharp corners,
    boxrule=0.8pt,
    left=8pt,
    right=8pt,
    top=8pt,
    bottom=8pt
]
You are given a sequence of keyframes of a video.
Your core objective is to evaluate a prediction of the next keyframe in the video.
When conducting your evaluation of the next keyframe prediction, you must focus on the following aspects:\par
\vspace{0.5em}
1. Style Consistency: Whether the image style of the next keyframe is consistent with former keyframes?
Answer an integer from 1 to 10, to indicate the consistency level, where 1 means totally inconsistent image style and 10 means totally the same image style.
Also output a concise justification of your answer in one or two sentences.\par
\vspace{0.5em}
2.1. Transformations - Objects: From the former keyframe to the next keyframe, whether all the appearance, disappearance or retention of objects accord with spatial-temporal causality and commonsense?
Answer ``yes'' or ``no''.
Also output a concise justification of your answer in one or two sentences.\par
\vspace{0.5em}
2.2. Transformations - Object States: From the former keyframe to the next keyframe, whether all the changes of objects' color, shape, texture or physical status accord with spatial-temporal causality and commonsense?
Ignore the changes caused mainly by image style shifts.
If there is no change of object states, answer ``unchange''.
Otherwise, answer ``yes'' or ``no'' according to the changes.
Also output a concise justification of your answer in one or two sentences.\par
\vspace{0.5em}
2.3. Transformations - Spatial Relations: From the former keyframe to the next keyframe, whether all the changes of objects' alignment, contact or positional relations accord with spatial-temporal causality and commonsense?
If there is no change of spatial relations, answer ``unchange''.
Otherwise, answer ``yes'' or ``no'' according to the changes.
Also output a concise justification of your answer in one or two sentences.\par
\vspace{0.5em}
2.4. Transformations - Motion or Actions: From the former keyframe to the next keyframe, whether all the changes or progressions of object motion or actions accord with spatial-temporal causality and commonsense?
If there is no change or progression of motion or actions, answer ``unchange''. Otherwise, answer ``yes'' or ``no'' according to the changes.
Also output a concise justification of your answer in one or two sentences.\par
\vspace{0.5em}
2.5. Transformations - Camera: From the former keyframe to the next keyframe, whether all the changes of camera viewpoint, zoom, focus or pan tilt accord with spatial-temporal causality and commonsense?
If there is no change of camera, answer ``unchange''.
Otherwise, answer ``yes'' or ``no'' according to the changes.
Also output a concise justification of your answer in one or two sentences.\par
\vspace{0.5em}     
2.6. Transformations - Background: From the former keyframe to the next keyframe, whether all the changes of background illumination, atmosphere, environment or setting accord with spatial-temporal causality and commonsense?
Ignore the changes caused mainly by image style shifts. If there is no change of background, answer ``unchange''.
Otherwise, answer ``yes'' or ``no'' according to the changes.
Also output a concise justification of your answer in one or two sentences.\par
\vspace{0.5em}
Final Output Format\par
\vspace{0.2em}
Conclude your evaluation with a structured JSON output following this exact schema:\par
\vspace{0.2em}
\{\par
``style\_consistency'':\{``rating'': integer, ``justification'': ``string''\},\par
``transformation\_objects'':\{``accordance'': ``yes'' or ``no'', ``justification'': ``string''\},\par
``transformation\_object\_states'':\{``accordance'': ``yes'', ``no'' or ``unchange'', ``justification'': ``string''\},\par
``transformation\_spatial\_relations'':\{``accordance'': ``yes'', ``no'' or ``unchange'', ``justification'': ``string''\},\par
``transformation\_motion\_actions'':\{``accordance'': ``yes'', ``no'' or ``unchange'', ``justification'': ``string''\},\par
``transformation\_camera'':\{``accordance'': ``yes'', ``no'' or ``unchange'', ``justification'': ``string''\},\par
``transformation\_background'':\{``accordance'': ``yes'', ``no'' or ``unchange'', ``justification'': ``string''\}\par
\}\par
\vspace{0.2em}
Do not output anything other than the final JSON as described.\par
\vspace{0.5em}
Here is the sequence of former video keyframes: $\langle$preceding state images before the state to be evaluated$\rangle$\par
\vspace{0.2em}
Here is the next keyframe prediction to evaluate: $\langle$the state image to be evaluated$\rangle$\par
\end{tcolorbox}

\section{Experimental Details}
\label{sec:exp_details}

\subsection{Baseline Methods}
\label{sec:baseline_methods}

\paragraph{Emu2}~\citep{sun2024generative} is a 37B-parameter generative multimodal foundation model designed to support both multimodal understanding and generation through in-context learning. It is trained on large-scale multimodal sequences, including text, image-text pairs, video-text pairs, and interleaved image-text-video data, using a unified autoregressive objective that predicts the next multimodal element, either a text token or a visual embedding. Architecturally, Emu2 consists of a visual encoder, a Transformer-based multimodal modeling module, and a visual decoder: images are encoded into continuous visual embeddings, interleaved with text tokens, modeled autoregressively, and then decoded back into images or videos when generation is required. 

\paragraph{BAGEL} is an open-sourced multimodal foundation model with 7B active parameters (14B total) trained on large-scale interleaved multimodal data, including text, image, video, and web data. 
It adopts a Mixture-of-Transformer-Experts (MoT) architecture that employs selective activation of modality-specific parameters and unifies multimodal understanding (i.e., understanding expert) and generation (i.e., generation expert) through shared self-attention operations. 
For visual understanding, it leverages a ViT~\citep{dosovitskiy2020image} encoder, initialized by SigLIP2-so400m/14~\citep{tschannen2025siglip} with a fixed 384-resolution, to convert the raw pixels into tokens. 
A two-layer MLP connector is adopted to match the feature dimension of the ViT tokens and the LLM hidden states.
For visual generation, it uses a pre-trained VAE model from FLUX~\citep{flux2024} to convert images from pixel space to latent space. The VAE encoder is frozen during training.

\paragraph{Story2Board} \citep{dinkevich2025story2board} is a training-free framework for expressive storyboard creation from natural language, with a special focus on enhancing visual consistency. 
It chooses FLUX.1-dev~\citep{flux2024}, a dense rectified flow transformer with 12B parameters, as the underlying VLM for storyboard generation. 
On top of FLUX.1-dev, it proposes two consistency-enhancing techniques: Latent Panel Anchoring, which preserves a shared character reference across panels, and Reciprocal Attention Value Mixing, which softly blends visual features between token pairs with strong reciprocal attention. 
For a fair comparison with other baselines, we always use the first ground-truth image of the storyboard as the reference panel at the top and the first narrative as the reference panel prompt, while the other transition narratives are panel prompts for different scenes.

\paragraph{ARLDM} \citep{pan2024synthesizing} trains a Stable Diffusion \citep{rombach2022high} module to auto-regressively generate each visual narrative image, which is conditioned on the BLIP \citep{li2022blip} embeddings of previous scenes' generated images and input textual constraints, and the CLIP \citep{radford2021learning} embedding of the next scene's input textual constraints.

\paragraph{MM-Interleaved} \citep{tian2024mm} trains a VLM, \textit{i.e.}, Vicuna \citep{zheng2023judging} with CLIP vision encoder, to model the interleaved sequence of previously generated images and their textual constraints, and a Stable Diffusion module to generate the next narrative image based on the output states of the VLM.
Both the VLM and the diffusion module are augmented by additional layers of cross-attention to sparse image features via Deformable Attention \citep{zhu2020deformable}.

\paragraph{StoryGen} \citep{liu2024intelligent} uses a dual-diffusion structure to perform the auto-regressive generation of narrative images.
It first adds noise to each previously generated image, and then the noisy image is de-noised by a Stable Diffusion module (conditioned on the image's corresponding input textual constraints), whose latent diffusion states are used as the extracted features of the image.
Conditioned on the next textual constraints and the concatenation of previous images' extracted features, a second Stable Diffusion module is trained to generate the next narrative image.

\subsection{Evaluation Benchmarks}
\label{sec:eval_benchmarks}

\paragraph{VinaBench}
VinaBench~\citep{gao2025vinabench} is a novel benchmark proposed for visual narrative generation, consisting of ~25K pairs of visual and textual narratives sampled from diverse visual storytelling datasets. 
VinaBench contains rich annotations about commonsense links between textual and visual narratives.
Moreover, VinaBench provides a set of global and scene-specific feature annotations to explicitly reveal the visual discourse. 
In this work, we primarily focus on the Visual Writing Prompts (VWP) subset that has 834 high-quality storyboards for downstream evaluations on visual narrative generation.
We convert part of the annotations into the form of JSON schema, which is then used in supervised fine-tuning as part of textual transition descriptions.
VinaBench proposes 5 metrics to measure the fine-grained alignment of each generated image with its corresponding scene's annotated narrative constraints: Non-Character Entities, Character Number, Character Attribute, Time of Day, and Location.
For each narrative sample, VinaBench adopts 3 metrics to assess the consistency of generated visual narrative images in terms of Style, Character, and Location. 
The prompt details can be found in Appendix~\ref{sec:vqa_prompts}.
We present the VinaBench evaluation results with GPT-4o as the judge in Table~\ref{tab:vinabench_score_main}, and further validate the evaluation results by using Gemini-2.5-Pro as the judge in Table~\ref{tab:vinabench_score_gemini}. 

\paragraph{LEGO}
LEGO~\citep{lai2024lego} introduces the egocentric action frame generation task, synthesizing an image depicting an action in the user's context (i.e., action frame) by conditioning on a user prompt and an input egocentric image. 
LEGO contains two egocentric action datasets, Ego4D and Epic-Kitchens, both of which are densely annotated with action starting time $t$ and ending time $\hat{t}$.
The LEGO dataset is constructed by selecting an egocentric image frame $\delta_i$ seconds before the action begins as the input $\mathcal{X}$, and an image $\delta_o$ seconds after the action begins as the target frame $\mathcal{Y}$.
After data filtering, LEGO ultimately contains 85,521/9,931 data samples for the train/test sets from Ego4D and 61,841/8,893 data samples for the train/test sets from Epic-Kitchens.
LEGO evaluates the action instruction-following performance in real-world simulation via 6 image-to-image similarity metrics.
It includes (1) EgoVLP score~\citep{lin2022egocentric}, (2) EgoVLP$^+$ score, and (3) CLIP score~\citep{radford2021learning}, all of which are based on contrastive learning. 
They also report (4) Fréchet Inception Distance (FID)~\citep{heusel2017gans}, (5) Peak Signal-to-Noise Ratio (PSNR), and (6) Learned Perceptual Image Patch Similarity (LPIPS)~\citep{zhang2018unreasonable}.
We report the performance of our model and baselines using these metrics in Table~\ref{tab:lego_instruction_following}.
In addition, we adopt two image-to-text metrics, BLIP-B(ase) and BLIP-L(arge), which utilize BLIP models to generate image captions of the output images and then calculate text-to-text similarity using the CLIP text encoder. We summarize results under image-to-text metrics in Table~\ref{tab:lego_image_to_text_score}.

\subsection{VLM Judge Prompts on VinaBench}
\label{sec:vqa_prompts}

In this section, we present the complete set of prompts used by the VLM judges on VinaBench, covering both per-scene alignment evaluation and cross-scene consistency evaluation.

\paragraph{Per-Scene Alignment.}
\begin{itemize}
  \item \textbf{Character Number.}
  At each timestamp, we provide the generated image together with the ground-truth character count from VinaBench annotations and ask the judge to identify the number of characters appearing in the image.
  The judge is required to respond using Arabic numerals only.

  \item \textbf{Character Attributes.}
  To assess fine-grained character fidelity, we include character--description pairs in the prompt and ask the judge whether the characters in the generated image match their corresponding descriptions.
  If all characters are consistent with the annotations, the judge answers ``Yes''; otherwise, it answers ``No''.

  \item \textbf{Non-character Entities.}
  Beyond characters, we evaluate non-character entities by asking whether the generated image contains or implies a specific target entity.
  The judge responds ``Yes'' or ``No'' for each entity, and we compute the proportion of positive responses.

  \item \textbf{Location.}
  We directly ask the judge whether the image is taken at a particular location specified by the ground-truth annotation.
  The judge responds ``Yes'' or ``No''.

  \item \textbf{Time.}
  The judge is prompted to determine whether the image corresponds to a specific time or time period described in the annotation, answering ``Yes'' or ``No''.
\end{itemize}

\paragraph{Cross-Scene Consistency.}
\begin{itemize}
  \item \textbf{Character.}
  Characters appearing at different timestamps are expected to maintain consistent visual characteristics.
  To evaluate this, we select images that should depict the same character according to the ground-truth annotations and ask the judge whether these images contain the same character with consistent attributes.
  This process is repeated for all characters, and we report the ratio of positive responses for each storyboard.

  \item \textbf{Location.}
  Similar to character consistency, we ask the judge whether multiple generated images are taken at the same location and report the corresponding positive response ratio for each storyboard.

  \item \textbf{Style.}
  We provide all generated images within a storyboard and ask the judge whether they share a consistent visual style.
  The judge responds ``Yes'' or ``No''.
\end{itemize}

\begin{tcolorbox}[
    colback=white,
    colframe=black!60,
    title=\textbf{VLM Judge Prompt for Character Number Alignment},
    coltitle=white,
    colbacktitle=black!60,
    fonttitle=\bfseries,
    sharp corners,
    boxrule=0.8pt,
    left=8pt,
    right=8pt,
    top=8pt,
    bottom=8pt
]
\textbf{\#\# System Prompt}\par
You are a helpful and objective judge. You must be strict, consistent, and unbiased, and rely only on observable evidence from the inputs.\par
\vspace{0.5em}
\textbf{\#\# Instruction}\par
\ph{image}\par
How many characters are in this image? Only answer an Arabic number.
\end{tcolorbox}

\begin{tcolorbox}[
    colback=white,
    colframe=black!60,
    title=\textbf{VLM Judge Prompt for Character Attributes Alignment},
    coltitle=white,
    colbacktitle=black!60,
    fonttitle=\bfseries,
    sharp corners,
    boxrule=0.8pt,
    left=8pt,
    right=8pt,
    top=8pt,
    bottom=8pt
]
\textbf{\#\# System Prompt}\par
You are a helpful and objective judge. You must be strict, consistent, and unbiased, and rely only on observable evidence from the inputs.\par
\vspace{0.5em}
\textbf{\#\# Instruction}\par
\ph{image}\par
\ph{First Character Name}: \ph{Description}\par
\ph{Second Character Name}: \ph{Description}\par
...\par
Do characters in this image fit into their descriptions? Only answer yes or no.\par
\end{tcolorbox}

\begin{tcolorbox}[
    colback=white,
    colframe=black!60,
    title=\textbf{VLM Judge Prompt for Non-Character Entities Alignment},
    coltitle=white,
    colbacktitle=black!60,
    fonttitle=\bfseries,
    sharp corners,
    boxrule=0.8pt,
    left=8pt,
    right=8pt,
    top=8pt,
    bottom=8pt
]
\textbf{\#\# System Prompt}\par
You are a helpful and objective judge. You must be strict, consistent, and unbiased, and rely only on observable evidence from the inputs.\par
\vspace{0.5em}
\textbf{\#\# Instruction}\par
\ph{image}\par
Does this image contain or imply \ph{entity \#1}? Only answer yes or no.\par
Does this image contain or imply \ph{entity \#2}? Only answer yes or no.\par
...\par
\end{tcolorbox}

\begin{tcolorbox}[
    colback=white,
    colframe=black!60,
    title=\textbf{VLM Judge Prompt for Location Alignment},
    coltitle=white,
    colbacktitle=black!60,
    fonttitle=\bfseries,
    sharp corners,
    boxrule=0.8pt,
    left=8pt,
    right=8pt,
    top=8pt,
    bottom=8pt
]
\textbf{\#\# System Prompt}\par
You are a helpful and objective judge. You must be strict, consistent, and unbiased, and rely only on observable evidence from the inputs.\par
\vspace{0.5em}
\textbf{\#\# Instruction}\par
\ph{image}\par
Is this image taken at a/an \ph{location}? Only answer yes or no.\par
\end{tcolorbox}

\begin{tcolorbox}[
    colback=white,
    colframe=black!60,
    title=\textbf{VLM Judge Prompt for Time Alignment},
    coltitle=white,
    colbacktitle=black!60,
    fonttitle=\bfseries,
    sharp corners,
    boxrule=0.8pt,
    left=8pt,
    right=8pt,
    top=8pt,
    bottom=8pt
]
\textbf{\#\# System Prompt}\par
You are a helpful and objective judge. You must be strict, consistent, and unbiased, and rely only on observable evidence from the inputs.\par
\vspace{0.5em}
\textbf{\#\# Instruction}\par
\ph{image}\par
Is this image taken at/in the \ph{time}? Only answer yes or no.\par
\end{tcolorbox}

\begin{tcolorbox}[
    colback=white,
    colframe=black!60,
    title=\textbf{VLM Judge Prompt for Character Consistency},
    coltitle=white,
    colbacktitle=black!60,
    fonttitle=\bfseries,
    sharp corners,
    boxrule=0.8pt,
    left=8pt,
    right=8pt,
    top=8pt,
    bottom=8pt
]
\textbf{\#\# System Prompt}\par
You are a helpful and objective judge. You must be strict, consistent, and unbiased, and rely only on observable evidence from the inputs.\par
\vspace{0.5em}
\textbf{\#\# Instruction}\par
\ph{Selected image \#1 that should contain the First Character}\par
\ph{Selected image \#2 that should contain the First Character}\par
...\par
Do all these images contain the same character \ph{First Character Name}: \ph{Description}? Only answer yes or no.\par
\vspace{1em}
\ph{Selected image \#1 that should contain the Second Character}\par
\ph{Selected image \#2 that should contain the Second Character}\par
...\par
Do all these images contain the same character \ph{Second Character Name}: \ph{Description}? Only answer yes or no.\par
...\par
\end{tcolorbox}

\begin{tcolorbox}[
    colback=white,
    colframe=black!60,
    title=\textbf{VLM Judge Prompt for Location Consistency},
    coltitle=white,
    colbacktitle=black!60,
    fonttitle=\bfseries,
    sharp corners,
    boxrule=0.8pt,
    left=8pt,
    right=8pt,
    top=8pt,
    bottom=8pt
]
\textbf{\#\# System Prompt}\par
You are a helpful and objective judge. You must be strict, consistent, and unbiased, and rely only on observable evidence from the inputs.\par
\vspace{0.5em}
\textbf{\#\# Instruction}\par
\ph{Selected image \#1 that should be taken at the First Location}\par
\ph{Selected image \#2 that should be taken at the First Location}\par
...\par
Are all these images taken at the same \ph{First Location}? Only answer yes or no.\par
\vspace{1em}
\ph{Selected image \#1 that should be taken at the Second Location}\par
\ph{Selected image \#2 that should be taken at the Second Location}\par
...\par
Are all these images taken at the same \ph{Second Location}? Only answer yes or no.\par
...\par
\end{tcolorbox}

\begin{tcolorbox}[
    colback=white,
    colframe=black!60,
    title=\textbf{VLM Judge Prompt for Style Consistency},
    coltitle=white,
    colbacktitle=black!60,
    fonttitle=\bfseries,
    sharp corners,
    boxrule=0.8pt,
    left=8pt,
    right=8pt,
    top=8pt,
    bottom=8pt
]
\textbf{\#\# System Prompt}\par
You are a helpful and objective judge. You must be strict, consistent, and unbiased, and rely only on observable evidence from the inputs.\par
\vspace{0.5em}
\textbf{\#\# Instruction}\par
\ph{image \#1}\par
\ph{image \#2}\par
\ph{image \#3}\par
...\par
Are all these images in the same style? Only answer yes or no.\par
\end{tcolorbox}

\subsection{Benchmark Examples}
\label{sec:benchmark_examples}

\begin{figure*}[t]
\centering
\includegraphics[width=1.0\linewidth]{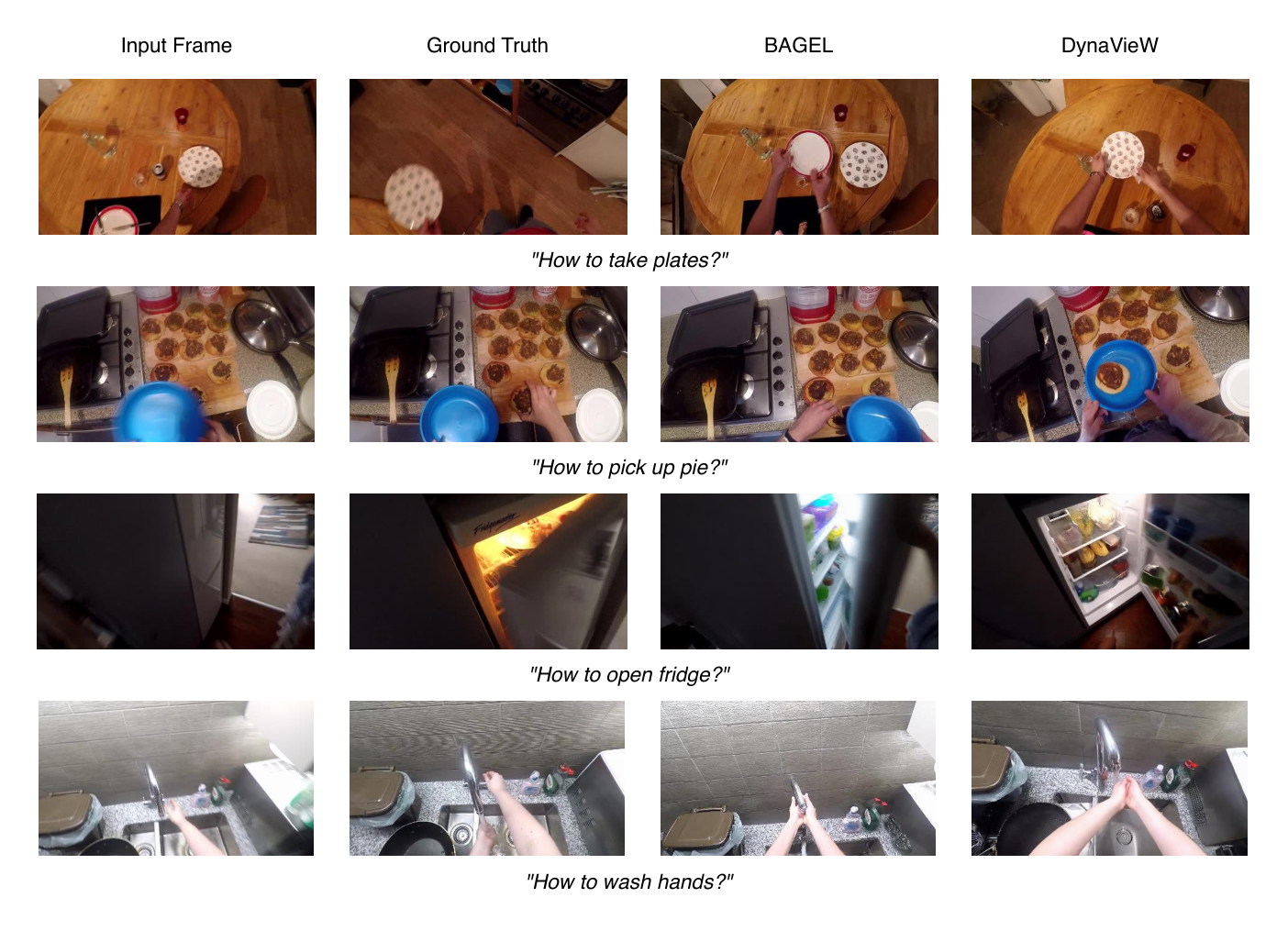}
\caption{World simulation outputs generated by \vwm{} and BAGEL after SFT, along with ground truth visual answers from LEGO.}
\label{fig:lego_examples}
\end{figure*}

In Figure~\ref{fig:storyboard_creation_examples}, we showcase several example storyboards created by our model, \vwm{} and BAGEL after SFT, along with ground-truth storyboards.
In Figure~\ref{fig:lego_examples}, we present 4 examples sampled from the Epic-Kitchen dataset.

\subsection{Supervised Fine-tuning Details}
\label{sec:sft_setting}
For supervised fine-tuning on each downstream benchmark (VinaBench or LEGO), we further fine-tune \vwm{} (and baseline models) on the official training samples provided by the benchmark, including $11652$ visual narrative samples from the Visual Writing Prompts (VWP) portion of VinaBench, or $147362$ world simulation samples from LEGO, respectively.
We fine-tune models on a cosine learning rate scheduler, with warm-up steps, total training steps, maximum learning rate and minimum learning rate set to $300$, $6000$, $5e^{-6}$ and $1e^{-7}$, respectively.
We deploy the training on $32$ NVIDIA GH200 GPUs, which takes about $2.5$ days.
Same as the pre-training, we freeze the weights of the VAE encoder during the fine-tuning of \vwm{}, and use AdamW \citep{loshchilov2018decoupled} optimizer with $\beta_{1}$, $\beta_{2}$ and $\epsilon$ set to $0.9$, $0.95$ and $1e^{-15}$, respectively.
The maximum norm for gradient clipping also remains to be $1.0$.

\subsection{Inference Settings}
\label{sec:inference_settings}

We systematically compare our model with baseline methods across multiple inference settings, including zero-shot, supervised fine-tuning, and model-assisted transition generation for the controllability experiment, where the transition format differs across settings.

\textbf{Zero-shot} Only the ground-truth transition narrative is included in the interleaved input stream.

\textbf{Supervised fine-tuning} In addition to the ground-truth transition narrative, we prompt both our model and the SFT baseline to generate a JSON schema in the style of VinaBench annotations, which is then added to the interleaved input stream.

\textbf{Model-assisted transition generation} In addition to the ground-truth transition narrative, we use Gemma-3 (27B-it) to generate a JSON schema matching the format of the pre-training data, which is then added to the interleaved input stream. For a fair comparison with BEGAL, we also convert the JSON schema into a natural-language representation as a variant. 


We admit that Gemma-3 (27B-it) has not been broadly pre-trained on world modelling tasks to understand the visual dynamics, but it has strong instruction-following performance. Therefore, we enhance its capability of visual dynamics understanding through sophisticated prompts, to fine-grainedly instruct its transition prediction.
Moreover, we expect Gemma to create diverse (more important than accurate in controllability experiments) transitions to test our DynaVieW model’s controllability of state simulation based on varied possible transitions, and noisy transitions can be used to examine the robustness (reflected by the low variance of scores shown in Table 3) of our model in resisting potentially infeasible state simulation conditions.

\begin{tcolorbox}[
    colback=white,
    colframe=black!60,
    title=\textbf{Zero-Shot Transition Template},
    coltitle=white,
    colbacktitle=black!60,
    fonttitle=\bfseries,
    sharp corners,
    boxrule=0.8pt,
    left=8pt,
    right=8pt,
    top=8pt,
    bottom=8pt
]
\ph{Textual Narrative}: Danny has come back home from his trip in Hawaii, and is greeting all of his neighbours.
\end{tcolorbox}

\begin{tcolorbox}[
    colback=white,
    colframe=black!60,
    title=\textbf{Supervised Fine-Tuning Transition Template (VinaBench)},
    coltitle=white,
    colbacktitle=black!60,
    fonttitle=\bfseries,
    sharp corners,
    boxrule=0.8pt,
    left=8pt,
    right=8pt,
    top=8pt,
    bottom=8pt
]
\ph{Textual Narrative}: Danny has come back home from his trip in Hawaii, and is greeting all of his neighbours.
\ph{Predicted Schema}:\par
\{\par
\quad "style": "photorealistic",\par
\quad "entities": [\par
\quad \quad "someone is arriving",\par
\quad \quad "someone",\par
\quad \quad "home",\par
\quad \quad "vacation"\par
\quad ],\par
\quad "characters": ["Danny", "neighbours"],\par
\quad "time": "afternoon",\par
\quad "location": "house",\par
\quad "caption": 
"..."\par
\}
\end{tcolorbox}

\begin{tcolorbox}[
    colback=white,
    colframe=black!60,
    title=\textbf{Gemma3-generated Natural Language Transition Template},
    coltitle=white,
    colbacktitle=black!60,
    fonttitle=\bfseries,
    sharp corners,
    boxrule=0.8pt,
    left=8pt,
    right=8pt,
    top=8pt,
    bottom=8pt
]
\ph{Textual Narrative}: Danny has come back home from his trip in Hawaii, and is greeting all of his neighbours.
\ph{Predicted Schema}:\par
High-level activity:\par
The overall activity depicted is Greeting Neighbors.\par
\vspace{1em}
Sub-activity: Person approaching and greeting\par
This sub-activity focuses on an individual moving toward nearby neighbors and initiating a greeting.\par
\vspace{1em}
Atomic action: Danny approaching neighbors\par
In this atomic action, Danny is introduced as an active participant in the scene. The surrounding environment remains unchanged and continues to include a house, street, plants, and a car, all of which persist throughout the action.\par
\vspace{1em}
Danny’s spatial position changes: 
he moves from being on or near the house (previously associated with the roof area) to a location closer to the neighbors. Although this movement is not directly visible in the current frame, it is inferred from the context. There are no changes in object states such as color, shape, physical condition, or texture, and no new physical contact or alignment relationships are formed.\par
\vspace{1em}
The action is characterized by initiating behaviors, specifically walking and greeting. Correspondingly, Danny exhibits translational motion as he walks toward the neighbors, while no rotational, oscillatory, or deformational motion is involved. The camera remains static, with no changes in viewpoint, zoom, pan, tilt, or focus. Likewise, the background—lighting, atmosphere, scene context, and overall environment—remains constant.\par
\vspace{1em}
Contribution and significance:\par
\quad •	At the atomic level, the introduction of Danny and his inferred movement toward the neighbors directly represents the act of approaching and greeting them.\par
\quad •	At the sub-activity level, this action fulfills the central purpose of the “Person approaching and greeting” sequence.\par
\quad •	Temporally, this atomic action serves as the initiating event, setting up the subsequent interaction between Danny and the neighbors.\par
\end{tcolorbox}

\begin{tcolorbox}[
    colback=white,
    colframe=black!60,
    title=\textbf{Gemma3-generated JSON-Schema Transition Template},
    coltitle=white,
    colbacktitle=black!60,
    fonttitle=\bfseries,
    sharp corners,
    boxrule=0.8pt,
    left=8pt,
    right=8pt,
    top=8pt,
    bottom=8pt
]
\ph{Textual Narrative}: Danny has come back home from his trip in Hawaii, and is greeting all of his neighbours.
\ph{Predicted Schema}:\par
\{\par
\quad "high\_level\_activity": "Greeting Neighbors",\par
\quad "sub\_activities": [
\{\par
\quad \quad \quad "name": "Person approaching and greeting",\par
\quad \quad \quad "atomic\_actions":
\{\par
\quad \quad \quad \quad \quad "name": "Danny approaching neighbors",\par
\quad \quad \quad \quad \quad "transitions": \{\par
\quad \quad \quad \quad \quad \quad "objects": \{\par
\quad \quad \quad \quad \quad \quad \quad "introduced": [\par
\quad \quad \quad \quad \quad \quad \quad \quad "Danny"\par
\quad \quad \quad \quad \quad \quad \quad ],\par
\quad \quad \quad \quad \quad \quad \quad "persistent": [\par
\quad \quad \quad \quad \quad \quad \quad \quad "House",\par
\quad \quad \quad \quad \quad \quad \quad \quad "Street",\par
\quad \quad \quad \quad \quad \quad \quad \quad "Plants",\par
\quad \quad \quad \quad \quad \quad \quad \quad "Car"\par
\quad \quad \quad \quad \quad \quad \quad ]\par
\quad \quad \quad \quad \quad \quad \},\par
\quad \quad \quad \quad \quad \quad "object\_states": \{...\}\par
\quad \quad \quad \quad \quad \quad "spatial\_relations": \{\par
\quad \quad \quad \quad \quad \quad \quad "positional\_changes": [\par
\quad \quad \quad \quad \quad \quad \quad \quad "Danny's position changes from being on the roof/near the house to presumably near the neighbors, though not visible in the current frame."\par
\quad \quad \quad \quad \quad \quad \quad ],\par
\quad \quad \quad \quad \quad \quad \quad ...\par
\quad \quad \quad \quad \quad \quad \},\par
\quad \quad \quad \quad \quad \quad "actions": \{\par
\quad \quad \quad \quad \quad \quad \quad "initiating": [\par
\quad \quad \quad \quad \quad \quad \quad \quad "Walking",\par
\quad \quad \quad \quad \quad \quad \quad \quad "Greeting"\par
\quad \quad \quad \quad \quad \quad \quad ],\par
\quad \quad \quad \quad \quad \quad \quad ... \par
\quad \quad \quad \quad \quad \quad \},\par
\quad \quad \quad \quad \quad \quad "motion": \{\par
\quad \quad \quad \quad \quad \quad \quad "translational": [\par
\quad \quad \quad \quad \quad \quad \quad \quad "Danny exhibits translational motion as he walks towards the neighbors (inferred)"\par
\quad \quad \quad \quad \quad \quad \quad ],\par
\quad \quad \quad \quad \quad \quad \quad ... \par
\quad \quad \quad \quad \quad \quad \},\par
\quad \quad \quad \quad \quad \quad "camera": \{...\}\par
\quad \quad \quad \quad \quad \quad "background": \{...\}\par
\quad \quad \quad \quad \quad \},\par
\quad \quad \quad \quad \quad "contributions": \{\par
\quad \quad \quad \quad \quad \quad "to\_atomic\_action": "The introduction of 'Danny' into the scene (though his initial position is not visible here) is directly linked to him starting to approach and greets his neighbors.",\par
\quad \quad \quad \quad \quad \quad "to\_sub\_activity": "This action fulfills the core goal of the 'Person approaching and greeting' sub-activity.",\par
\quad \quad \quad \quad \quad \quad "temporal\_significance": "This is the initiating action that sets the stage for showcasing the arrival and subsequent interaction with neighbors." \par
                    \}
                \}
            ]
        \}
    ]
\}

\end{tcolorbox}
\begin{table*}[t]
\centering
\caption{
Evaluation results of visual narrative generation on VinaBench, with Gemini-2.5-Pro as the judge.
``Non-Char. Ent.'', ``Char. Num.'' and ``Char. Attr.'' indicate Non-Character Entities, Character Number and Character Attribute, respectively.
The best results within each block are \textbf{in bold} and the second-best are \underline{underlined}.
}
\resizebox{1.0\textwidth}{!}{
\smallskip
\begin{tabular}{lccccccccc}
\toprule
\multirow{2}*{\textbf{Model}}
& \multicolumn{5}{c}{\textbf{Per-Scene Alignment}} 
& \multicolumn{3}{c}{\textbf{Cross-Scene Consistency}} 
& \multirow{2}*{\textbf{Average}}
\\
\cmidrule(lr){2-6} \cmidrule(lr){7-9}
& Non-Char. Ent. & Char. Num. & Char. Attr. & Time & Location
& Style & Character & Location & 
\\
\midrule
\multicolumn{10}{c}{\textbf{Zero-Shot Generalization Performance}} \\
\midrule
BAGEL
& \underline{0.720} & \underline{0.280} & \textbf{0.533} & \underline{0.626} & \underline{0.537} 
& 0.557 & 0.112 & \underline{0.323} 
& 0.435 \\

Story2Board
& \textbf{0.723} & \textbf{0.339} & 0.428 & 0.369 & 0.411 
& \textbf{0.815} & \underline{0.280} & 0.225 
& \underline{0.447} \\

\rowcolor{gray!20}
DynaVieW
& 0.719 & 0.278 & \underline{0.527} & \textbf{0.655} & \textbf{0.556} 
& \underline{0.645} & \textbf{0.284} & \textbf{0.610} 
& \textbf{0.530} \\

\midrule
\multicolumn{10}{c}{\textbf{Supervised Fine-Tuning}} \\
\midrule

ARLDM
& 0.765 & 0.360 & 0.576 & 0.469 & 0.465
& 0.397 & 0.061 & 0.097
& 0.356 \\

MM-Interleaved
& \underline{0.770} & \underline{0.375} & 0.581 & 0.521 & 0.499
& \textbf{0.650} & \underline{0.081} & 0.152
& \underline{0.422} \\

StoryGen
& 0.738 & 0.370 & 0.440 & 0.367 & 0.412
& 0.131 & 0.048 & 0.055
& 0.272 \\

BAGEL
& 0.760 & \textbf{0.392} & \textbf{0.585} & \underline{0.575} & \textbf{0.547}
& 0.569 & 0.058 & \underline{0.168}
& 0.418 \\

\rowcolor{gray!20}
DynaVieW
& \textbf{0.777} & 0.371 & \underline{0.583} & \textbf{0.601} & \underline{0.541}
& \underline{0.643} & \textbf{0.082} & \textbf{0.239}
& \textbf{0.448} \\

\bottomrule
\end{tabular}
}
\label{tab:vinabench_score_gemini}
\end{table*}

\begin{table*}[t]
\centering
\caption{
Image-to-text metrics of our model and baselines on LEGO.
The best results within each block are \textbf{in bold}.
}
\resizebox{0.45\linewidth}{!}{
\smallskip
\begin{tabular}{lcccc}
\toprule
\multirow{2}*{\textbf{Model}}
& \multicolumn{2}{c}{\textbf{Ego4D}} 
& \multicolumn{2}{c}{\textbf{Epic-Kitchens}} 
\\
\cmidrule(lr){2-3} \cmidrule(lr){4-5}
& BLIP-B & BLIP-L 
& BLIP-B & BLIP-L 
\\
\midrule
LEGO
& 20.38 & 20.70 
& 26.98 & 27.41 \\

BAGEL
& 23.01 & 22.74 
& 26.02 & 26.60 \\

\rowcolor{gray!20}
DynaVieW
& \textbf{25.02} & \textbf{24.90}
& \textbf{27.27} & \textbf{27.93} \\

\bottomrule
\end{tabular}
}
\label{tab:lego_image_to_text_score}
\end{table*}

\section{More Experimental Results}
\label{sec:additional_exp_results}


\subsection{The Trade-off Between Per-Scene Alignment and Cross-Scene Consistency}
\begin{figure*}[htbp]
\centering
    \begin{subfigure}[t]{\linewidth}
        \centering
        \includegraphics[width=1.0\linewidth]{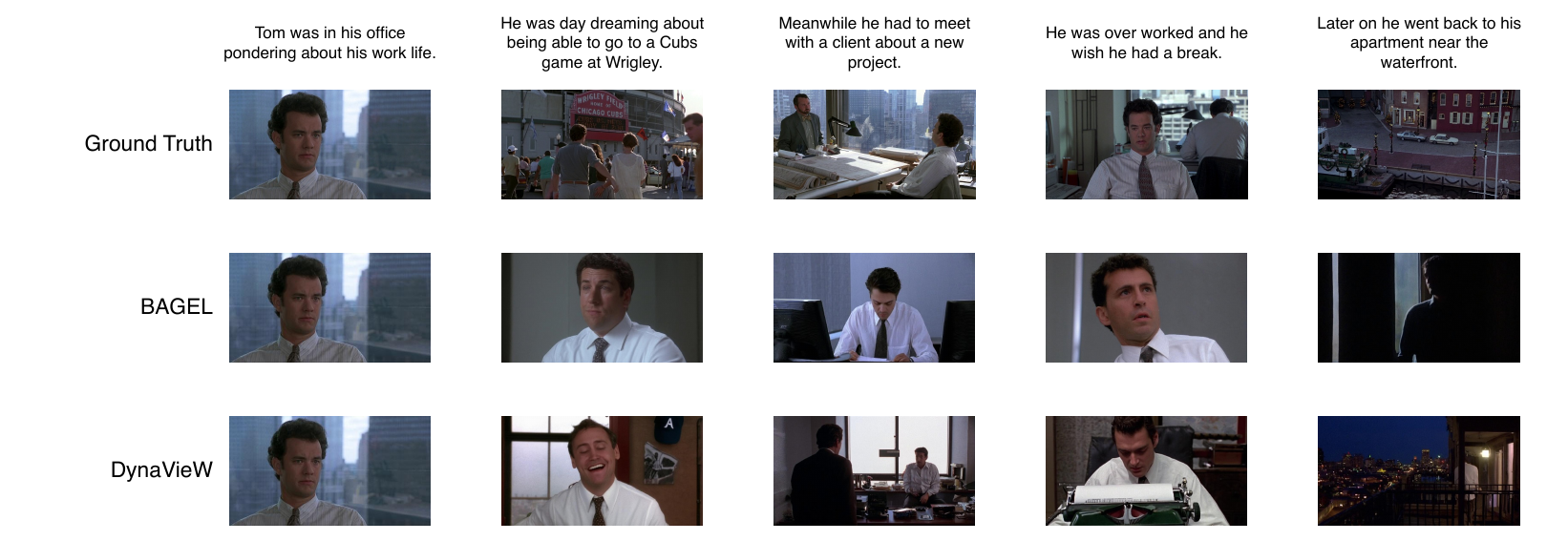}
        \caption{An exhausted office worker daydreams of escape before returning home after a demanding day.}
        \label{fig:first_storyboard_example}
    \end{subfigure}
    
    \vspace{1em}
    
    \begin{subfigure}[t]{\linewidth}
        \centering
        \includegraphics[width=1.0\linewidth]{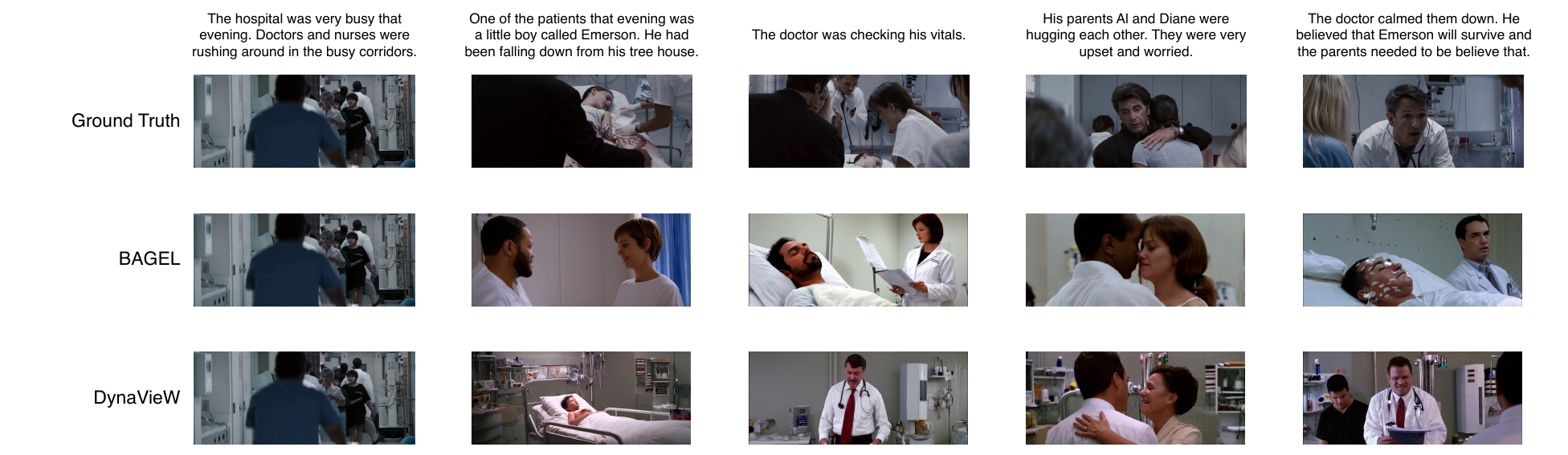}
        \caption{Doctors treat an injured child while his distressed parents await reassurance.}
        \label{fig:second_storyboard_example}
    \end{subfigure}
    
    \vspace{1em}
    
    \begin{subfigure}[t]{\linewidth}
        \centering
        \includegraphics[width=1.0\linewidth]{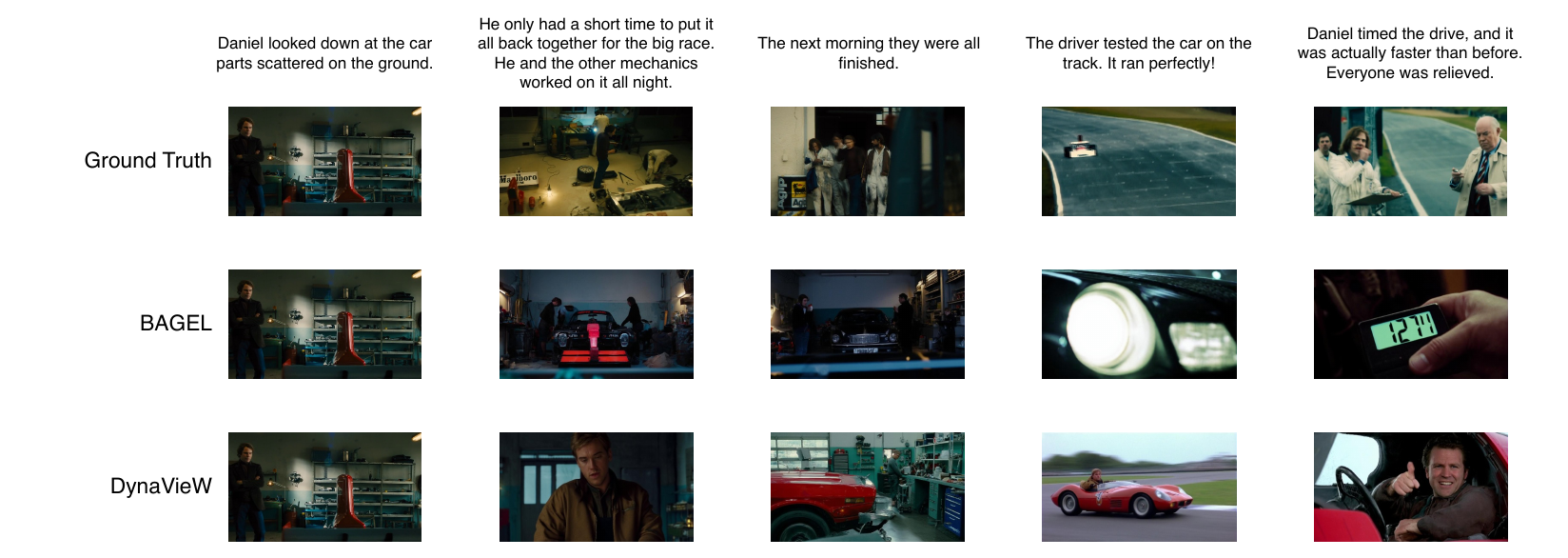}
        \caption{A race car is rebuilt overnight and proves its improved performance during testing.}
        \label{fig:third_storyboard_example}
    \end{subfigure}
\caption{Example storyboards generated by \vwm{} and BAGEL after SFT, along with ground truth storyboards from VinaBench.}
\label{fig:storyboard_creation_examples}
\end{figure*}

The experimental results in Table~\ref{tab:vinabench_score_main} indicate that supervised fine-tuning on downstream data improves per-scene alignment (0.566 vs 0.499), but at the cost of reduced cross-scene consistency (0.654 vs 0.761). We attribute this trade-off to the tendency of supervised fine-tuning to encourage close imitation of ground-truth scenes, including detailed transitions and visual manifestations, which leads to higher per-scene alignment scores. However, as illustrated by the storyboard examples in Figure~\ref{fig:storyboard_creation_examples}, this overemphasis on per-scene alignment promotes richer local transformations while weakening the visual coherence between consecutive scenes, ultimately degrading long-term consistency.

\subsection{Gemini-2.5-Pro As the Judge}
\label{sec:gemini_2.5_pro_eval_results}

Table~\ref{tab:vinabench_score_gemini} compares visual narrative generation performance under both zero-shot and supervised fine-tuning settings using Gemini-2.5-Pro as the judge. 
In the zero-shot regime, DynaVieW achieves the highest overall average score (0.530), substantially outperforming BAGEL (0.435) and Story2Board (0.447).
Under supervised fine-tuning, DynaVieW continues to demonstrate robust performance, achieving the highest average score (0.448) among all methods.
While BAGEL attains slightly better per-scene alignment scores on character number, attributes, and location, DynaVieW consistently outperforms alternatives on cross-scene consistency metrics, particularly for character (0.082) and location (0.239).
Overall, the additional evaluations with Gemini-2.5-Pro as the judge further demonstrate our model’s strong world-modeling capability.

\subsection{Image-to-Text Metrics on LEGO}
\label{sec:lego_image_to_text_score}

We report results of our model and baselines under six image-to-image metrics on LEGO in Table~\ref{tab:lego_instruction_following}, which are based on comparisons between the output images and ground-truth images. 
Prior work~\citep{lai2024lego} shows that the widely-used image-to-text CLIP score cannot align actions with egocentric images due to the domain gap.
So we implement another two image-to-text metrics, BLIP-B and BLIP-L, where we generate captions for the output images using two vision-language pre-trained BLIP~\citep{li2022blip} models and calculate the text-to-text similarity score between the captions and action instructions.
We report the results of BLIP-B and BLIP-L in Table~\ref{tab:lego_image_to_text_score}.

\subsection{Compute Cost and Latency of \vwm{}}
\label{sec:latency_cost}
We have tested the compute and latency costs of models on the downstream visual narrative generation task using a single NVIDIA H100 GPU, averaging results over 300 random testing samples from VinaBench dataset. 
The latency (in seconds) and the approximated KV-Cache (in MB) consumed for generating a visual narrative state with respect to the number of its preceding (previously generated) states in the context is shown in Table~\ref{tab:latency_compute_cost}. \vwm{} and BAGEL do state-only predictions.

We acknowledge that the maximum length of visual sequences that can be handled by our \vwm{} model is limited to the backbone model (BAGEL) capacity and computing, which only support modeling a maximum number of 36,864 tokens. And the evaluation on VinaBench, which only consists of short storyboards (e.g., 5-9 visual states), is constrained.
Scaling up to longer sequences requires a larger token budget and is out of our paper scope, which we leave for future work.

\begin{table}[htbp]
\centering
\caption{Latency and compute cost of \vwm{} and BAGEL under different numbers of preceding states}
\label{tab:latency_compute_cost}
\resizebox{0.75\textwidth}{!}{
\begin{tabular}{c cc cc}
\toprule
\multirow{2}{*}{\# Preceding States} 
& \multicolumn{2}{c}{Latency (s)} 
& \multicolumn{2}{c}{Compute Cost (MB)} \\
\cmidrule(lr){2-3} \cmidrule(lr){4-5}
& DynaVieW & BAGEL & DynaVieW & BAGEL \\
\midrule
1 & $17.92 \pm 0.77$ & $18.15 \pm 0.76$ & $407.71 \pm 19.71$ & $406.77 \pm 20.66$ \\
2 & $18.50 \pm 0.89$ & $18.74 \pm 0.85$ & $502.91 \pm 28.75$ & $501.38 \pm 30.10$ \\
3 & $19.11 \pm 1.01$ & $19.32 \pm 1.00$ & $582.73 \pm 35.10$ & $580.32 \pm 35.30$ \\
4 & $19.72 \pm 1.12$ & $19.96 \pm 1.10$ & $646.08 \pm 49.59$ & $645.85 \pm 48.35$ \\
5 & $20.30 \pm 1.22$ & $20.55 \pm 1.22$ & $724.30 \pm 61.51$ & $722.40 \pm 62.08$ \\
\bottomrule
\end{tabular}
}
\end{table}

\subsection{More Baselines on LEGO}
\label{sec: more_baselines}
We add more baselines from prior work~\citep{lai2024lego} on LEGO benchmark. The evaluation results in Table~\ref{tab:more_lego_baselines} (tested on the Epic-Kitchen portion) show that the performance of \vwm{} is consistently better than other (SFT) baselines.

\begin{table}[htbp]
\centering
\caption{More Baselines on LEGO Benchmark
}
\resizebox{0.8\textwidth}{!}{
\smallskip
\begin{tabular}{@{~}lc@{~~~}c@{~~~}c@{~~~}c@{~~~}c@{~~~}c@{~}}
\toprule
& FID $\downarrow$
& PSNR $\uparrow$
& LPIPS $\downarrow$
& CLIP $\uparrow$
& EgoVLP $\uparrow$
& EgoVLP$^+$ $\uparrow$ \\
\midrule

ProxEdit~\citep{han2023improving}
& 51.35 & 11.06 & 46.35 & 65.80 & 32.27 & 52.77 \\
SDEdit~\citep{meng2021sdedit}
& 27.41 & 11.30 & 43.33 & 74.76 & 33.84 & 56.80 \\

IP2P~\citep{brooks2023instructpix2pix}
& 20.64 & 11.23 & 40.82 & 77.03 & 42.97 & 61.06 \\

LEGO~\citep{lai2024lego}
& 21.57 & 11.33 & 40.36 & 78.63 & 45.89 & \textbf{62.66} \\

\rowcolor{gray!20}
\vwm{} (Zero-Shot)
& 18.33 & \textbf{11.48} & 41.78 & 77.79 & 43.60 & 58.83 \\

\rowcolor{gray!20}
\vwm{} (SFT)
& \textbf{10.31} & 11.31 & \textbf{39.20} & \textbf{84.19} & \textbf{48.72} & 61.77 \\

\bottomrule
\end{tabular}
}
\label{tab:more_lego_baselines}
\end{table}

\end{document}